\documentclass[lettersize,journal]{IEEEtran}
\usepackage{amsmath,amsfonts}
\usepackage{algorithmic}
\usepackage{array}
\usepackage[caption=false,font=normalsize,labelfont=sf,textfont=sf]{subfig}
\usepackage{textcomp}
\usepackage{stfloats}
\usepackage{url}
\usepackage{verbatim}
\usepackage{graphicx}

\usepackage{color}
\usepackage{multirow}
\usepackage{array} 
\usepackage{svg}
\usepackage{wrapfig}
\usepackage{booktabs}       
\usepackage[linesnumbered,boxed,ruled,commentsnumbered]{algorithm2e}
\def\BibTeX{{\rm B\kern-.05em{\sc i\kern-.025em b}\kern-.08em
    T\kern-.1667em\lower.7ex\hbox{E}\kern-.125emX}}
\usepackage{balance}
\begin{document}
\title{Learning to Generate Parameters of ConvNets for Unseen Image Data}
\author{Shiye~Wang,
        Kaituo~Feng,
        Changsheng~Li,
        Ye~Yuan,
        Guoren~Wang
\IEEEcompsocitemizethanks{
\IEEEcompsocthanksitem 
Shiye Wang, Kaituo Feng, Changsheng Li, Ye Yuan, and Guoren Wang are with the school of computer science and technology, Beijing Institute of Technology, Beijing, China. E-mail: \{sywang@bit.edu.cn; kaituofeng@gmail.com; lcs@bit.edu.cn; yuan-ye@bit.edu.cn; wanggrbit@126.com.\}
\IEEEcompsocthanksitem Shiye Wang and Kaituo Feng are equally contributed.
\IEEEcompsocthanksitem Corresponding author: Changsheng Li.
}}


\markboth{IEEE Transactions on Image Processing}%
{How to Use the IEEEtran \LaTeX \ Templates}

\maketitle

\begin{abstract}
Typical Convolutional Neural Networks (ConvNets) depend heavily on large amounts of image data and resort to an iterative optimization algorithm (e.g., SGD or Adam) to learn network parameters, making training very time- and resource-intensive. In this paper, we propose a new training paradigm and formulate the parameter learning of ConvNets into a  prediction task:  
considering that there exist correlations between image datasets and their corresponding optimal network parameters of a given ConvNet, we explore if we can learn a hyper-mapping between them to capture the relations,
such that we can directly predict the parameters of the network for an image dataset never seen during the training phase. To do this, we put forward a new hypernetwork-based model, called PudNet, which intends to learn a mapping between datasets and their corresponding network parameters, then predicts parameters for unseen data with only a single forward  propagation.
Moreover, our model benefits from 
a series of adaptive hyper-recurrent units sharing weights to capture the dependencies of parameters among different network layers.
Extensive experiments demonstrate that our proposed method achieves good efficacy  for unseen image datasets in two kinds of settings: Intra-dataset prediction  and Inter-dataset prediction. 
Our PudNet can also well scale up to large-scale datasets, e.g., ImageNet-1K.
It takes 8,967 GPU seconds to train ResNet-18 on the ImageNet-1K using GC from scratch and obtain a top-5 accuracy of 44.65\%. However, our PudNet costs only 3.89 GPU seconds to predict the network parameters of ResNet-18 achieving comparable performance (44.92\%), more than 2,300 times faster than the traditional training paradigm.

\end{abstract}

\begin{IEEEkeywords}
Parameter generation, hypernetwork, adaptive hyper-recurrent units.
\end{IEEEkeywords}

\section{Introduction}
Convolutional Neural Networks (ConvNets)  have yielded superior performance in a variety of fields in the past decade, such as computer vision \cite{kendall2017uncertainties}, reinforcement learning \cite{zheng2018drn,fujimoto2018addressing}, etc. 
One of the keys to success for ConvNets stems from huge amounts of image training data. To optimize ConvNets, the traditional training paradigm  takes advantage of an iterative optimization algorithm (e.g., SGD) to train the model in a mini-batch manner, leading to huge time and resource consumption.  For example, when training ResNet-101~\cite{he2016deep} on ImageNet~\cite{deng2009ImageNet}, it often takes several days or weeks for the model to be well optimized with GPU involved.
Thus, how to accelerate the training process of ConvNets is an emergent topic in deep learning. 

Nowadays, many methods for accelerating the training of deep neural networks have been proposed \cite{DBLP:journals/corr/KingmaB14,ioffe2015batch,chen2018bi}. The representative works include optimization-based techniques by improving the stochastic gradient descent \cite{DBLP:journals/corr/KingmaB14,yong2020gradient,anil2020scalable}, 
normalization based techniques \cite{ioffe2015batch,salimans2016weight,ba2016layer}, parallel training techniques \cite{chen2018bi,kim2019parallax}, et. Although these methods have shown promising potential to speed up network training, they still adhere to the traditional iterative-based training paradigm. 

In this paper, we investigate a new training paradigm for ConvNets. In contrast to previous works accelerating the training of the network, we formulate the parameter training problem into a prediction task: given a ConvNet architecture,  we attempt to learn a hyper-mapping between image datasets and their corresponding optimal network parameters, and then leverage the hyper-mapping to directly predict the network parameters for a new image dataset unseen during training. A basic assumption behind the above prediction task is the presence of correlations between image datasets and their corresponding parameters of a given ConvNet.


\begin{figure}[!t]
\centering
\includegraphics[width=0.65\columnwidth]{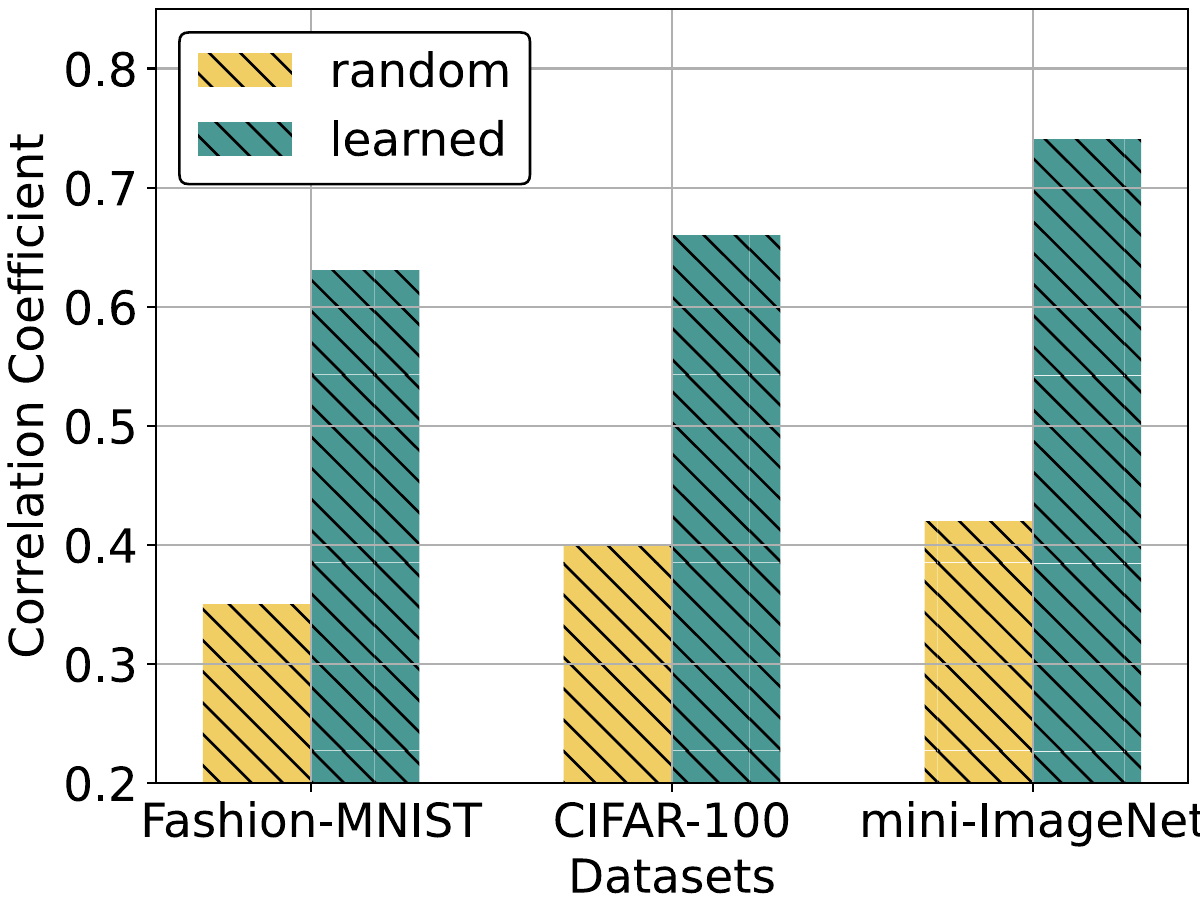}
\vspace{-0.1in}
\caption{Correlation coefficients between training datasets and the network parameters on the Fashion-MNIST \cite{xiao2017fashion}, CIFAR-100 \cite{krizhevsky2009learning}, Mini-ImageNet \cite{vinyals2016matching} datasets, respectively. `learned’ depicts correlations between training datasets and the corresponding optimal network parameters. `random’ denotes correlations between training datasets and the network parameters  selected  randomly from 1000 groups.}
\label{correlation}
\vspace{-0.2in}
\end{figure}

To demonstrate the rationality of this assumption,  we perform the following experiment: 
for an image  dataset, we first randomly sample 3000 images to train a 3-layer convolutional neural  network until convergence. Then we conduct the average pooling operation to the original inputs as a vector representation of the training data. We repeat the above experiment 1000 times and thus obtain 1000 groups of representations and the corresponding network parameters. Finally, we utilize Canonical Correlation Analysis (CCA) \cite{weenink2003canonical} to assess the correlations between the training data and the network parameters by the above 1000 data groups. 
Fig. \ref{correlation} shows the results, which clearly illustrate the presence of correlations between the training datasets and their network parameters for a given network architecture.

In light of this, we propose PudNet, a new hypernetwork-based model, to learn a hyper-mapping between image datasets and ConvNet parameters. 
PudNet first summarizes the characteristics of image datasets by compressing them into different vectors as their sketches. Then, it extends the traditional hypernetwork \cite{DBLP:conf/iclr/HaDL17} to predict network parameters for different layers based on these vectors. 
Considering the interdependency of model parameters across layers, 
we design an adaptive hyper-recurrent unit (AHRU) sharing weights to capture the relations among them.
This design enables the predicted parameters of each layer to adapt to both dataset-specific information and the predicted parameter information from previous layers, thereby enhancing the performance of PudNet.
Finally, it is worth noting that training PudNet becomes prohibitive if thousands of datasets need to be prepared and the network of a given architecture has to be trained on each of these datasets to obtain their respective optimal parameters.
Instead, we adopt a meta-learning based approach~\cite{finn2017model} to train PudNet. 

In the experiment, our PudNet showcases the remarkable efficacy  for unseen yet related image datasets. 
For example, it takes around 54, 119, and 140 GPU seconds to train ResNet-18 using Adam from scratch and obtain top-1 accuracies of 99.91\%, 74.56\%, and 71.84\%  on the Fashion-MNIST~\cite{xiao2017fashion}, CIFAR-100~\cite{krizhevsky2009learning}, and Mini-ImageNet~\cite{vinyals2016matching} datasets respectively.
In contrast, our method costs only around 0.5 GPU seconds to predict the parameters of ResNet-18 while still achieving top-1 accuracies of 96.24\%, 73.33\%, and 71.57\% on the same three datasets respectively.
This represents a substantial improvement as our method is at least 100 times faster than the traditional training paradigm.
More surprisingly, PudNet exhibits impressive performance even on cross-domain unseen image datasets, demonstrating the good generalization ability of PudNet.

Our contributions are summarized as follows: 
\begin{itemize}
\item 
Considering the correlations between image datasets and their corresponding parameters of a given ConvNet, we propose a general training paradigm for ConvNets by  formulating network training into a prediction task.
\item We extend hypernetwork to learn the correlations between image datasets and the corresponding ConvNet parameters, enabling the direct generation of parameters for unseen image data with only a single forward propagation.
Furthermore, a theoretical analysis of the hyper-mapping is provided to offer interpretability.
\item We design an adaptive hyper-recurrent unit that empowers the predicted parameters of each layer to encompass both the dataset-specific information and the inter-layer parameter dependencies.
\item  We perform extensive experiments on image datasets in terms of both Intra-dataset and Inter-dataset prediction tasks, demonstrating the efficacy of our method.
We expect that these results will motivate more researchers to explore this research direction. 
\end{itemize}

\section{Related work}
\subsection{Hypernetwork}
Hypernetwork  in \cite{DBLP:conf/iclr/HaDL17} aims to decrease the number of training parameters, by training a  smaller network to generate the parameters of  a larger network on a fixed dataset. 
Hypernetwork has been gradually applied to various tasks \cite{krueger2017bayesian,DBLP:conf/iclr/ZhangRU19,DBLP:conf/iclr/OswaldHSG20,li2020dhp,shamsian2021personalized, yin2022sylph,dinh2022hyperinverter,alaluf2022hyperstyle}. 
\cite{DBLP:conf/iclr/OswaldHSG20} proposes a task-conditioned hypernetwork to overcome catastrophic forgetting in continual learning. 
Bayesian hypernetwork \cite{krueger2017bayesian}  is proposed to approximate Bayesian inference in neural networks.
\cite{DBLP:conf/acl/MahabadiR0H20} and \cite{liu2022polyhistor} utilize multiple layer-wise hypernetworks shared across different tasks to generate weights for the adapters \cite{houlsby2019parameter} in different layers.
GHN-2 proposed in \cite{knyazev2021parameter} attempts  to build a mapping between the network architectures and  network parameters, where the dataset is always fixed.
HyperTransformer \cite{zhmoginov2022hypertransformer} adopts a {transformer-based hypernetwork} to blend the information of a small support set to generate parameters for a small CNN in the few-shot setting. However, due to the high time complexity of the attention operation, this approach faces the scalability issue as networks become deeper  and the support set {becomes} larger.
Different from these works, we aim to directly predict the parameters of a given ConvNet for unseen image datasets.

\subsection{Acceleration of Network Training}
Many works have been proposed to speed up the training  of deep neural networks, including optimization-based methods \cite{duchi2011adaptive,DBLP:journals/corr/KingmaB14,yong2020gradient}, normalization based methods \cite{ioffe2015batch,ba2016layer,salimans2016weight}, parallel training methods \cite{chen2018bi,kim2019parallax, fei2021efficient}, etc.
Optimization-based methods mainly aim to improve the stochastic gradient descent. 
For instance, Adaptive Moment Estimation (Adam) \cite{DBLP:journals/corr/KingmaB14} achieves this by combining the strengths of AdaGrad \cite{duchi2011adaptive} and RMSProp \cite{tieleman2017divide}.
By comprehensively considering both the first-order moment estimation and the second-order moment estimation of the gradients, Adam \cite{DBLP:journals/corr/KingmaB14} enables the utilization of more efficient and adaptive learning rates for each parameter.
\cite{yong2020gradient} proposes a gradient centralization method that centralizes gradient vectors to improve the Lipschitzness of the loss function.
Normalization based methods intend to design  good normalization methods to speed up  training. 
Batch normalization \cite{ioffe2015batch} is a prominent work on this line, known for its ability to smooth the optimization landscape and achieve fast convergence \cite{santurkar2018does}.
Parallel training methods \cite{fei2021efficient,sapio2021scaling} commonly leverage multiple hardware resources to conduct  parallel training, thereby reducing training time by distributing computational tasks among distributed devices.
However, these methods still follow the traditional iterative based training paradigm. Different from them, we attempt to explore a new training paradigm and transform the network training problem into a parameter prediction task. 

\subsection{Meta-Learning}
Meta-learning aims to learn a model from a variety of tasks, such that it can quickly adapt to a new learning task using only a small number of training samples \cite{hospedales2021meta}.
So far, many meta-learning methods have been proposed \cite{yao2020automated,snell2017prototypical,hospedales2021meta,santoro2016meta, munkhdalai2017meta}.
The representative algorithms include optimization-based methods and metric-based  methods \cite{hospedales2021meta}.
Optimization-based meta-learning methods usually train the model for easily finetuning by a small number of gradient steps \cite{finn2017model, DBLP:conf/icml/JiangK022}. 
Metric-based meta-learning methods attempt to learn to match the training sets with the class centroids and predict the label of training sets with the matched classes \cite{snell2017prototypical,chen2021meta,xie2022joint}.
Because of its powerful ability, meta-learning has been widely used for various tasks, including few-shot learning \cite{vinyals2016matching,snell2017prototypical,hu2022pushing,xie2022joint},  zero-shot learning \cite{nooralahzadeh2020zero}, etc.
Different from the above methods, we propose a new training paradigm of neural networks, where we attempt to directly predict network parameters for an unseen dataset  with only a single forward propagation without training on the dataset.

\begin{figure*}
\begin{center}
\centerline{\includegraphics[width=0.8\textwidth]{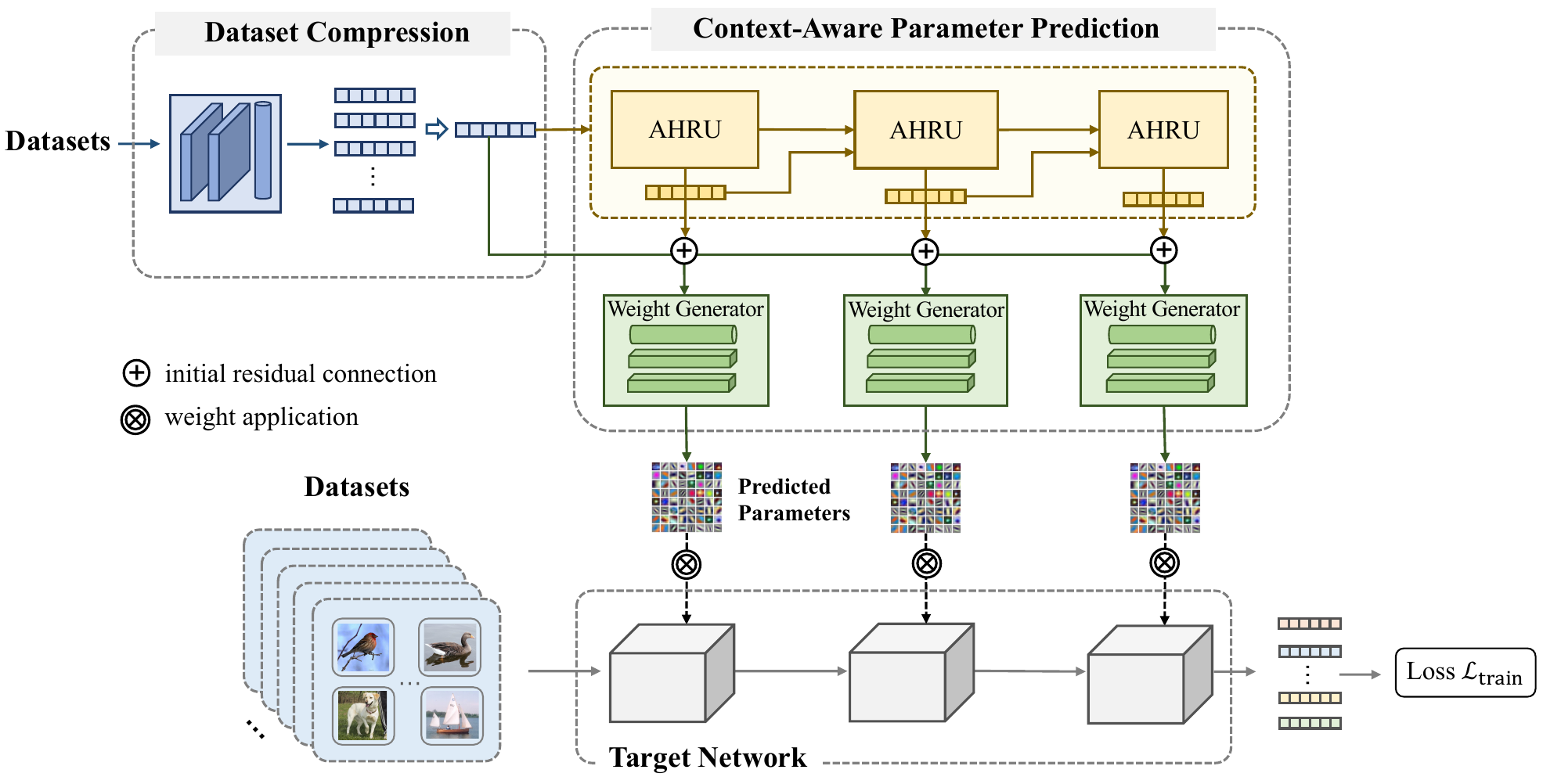}}
\vspace{-0.1in}
\caption{Overview of our PudNet. PudNet compresses each image dataset into a sketch with a fixed size and utilizes the designed hypernetwork to generate parameters of a target ConvNet using this sketch.
Specifically, PudNet exploits the adaptive hyper-recurrent units (AHRU) to process the dataset sketch, capturing the dependencies of parameters among different network layers for predicting the parameter representations. Subsequently, the weight generators are employed to produce the dataset-specific parameters of the target ConvNet.}
\label{frame}
\end{center}
\vspace{-0.2in}
\end{figure*}

\section{Proposed Method}
\subsection{Preliminaries and Problem Formulation}
We denote $H_\theta$ as our hypernetwork parameterized by $\theta$. Let $\mathcal{D}^{train}=\{ D_i \}_{i=1}^{\mathcal{N}}$ be the set of image training datasets, where $D_i$ is the $i^{th}$ image dataset and $\mathcal{N}$ is the number of image training datasets. Each sample $x_j \in D_i$ has a label $y_j \in \mathcal{C}^{tr}_i$, where $\mathcal{C}^{tr}_i$ is the class set of $D_i$.
We use $\mathcal{C}^{tr}= \bigcup_{i=1}^\mathcal{N}\mathcal{C}^{tr}_i$ to denote the whole label set of image training datasets.
Similarly, we define $\mathcal{D}^{test}$ as the set of unseen image datasets used for testing and  $\mathcal{C}^{te}$ as the set containing all labels in $\mathcal{D}^{test}$. 

In contrast to the traditional training paradigm, we attempt to explore a new training paradigm and formulate the  training of ConvNets into a parameter prediction task. To this end, we propose the following objective function:
\begin{equation}
\label{obj1}
\setlength{\abovedisplayskip}{0.05in}
\setlength{\belowdisplayskip}{0.05in}
\mathop{\arg\min}\limits_{\theta} \quad \sum_{i=1}^\mathcal{N}\mathcal{L}(\mathcal{F}(D_i,\Omega;H_{\theta}),\mathcal{M}_i^\Omega),
\end{equation}
where $\mathcal{F}(D_i,\Omega; H_{\theta})$ denotes a forward propagation of our hypernetwork $H_{\theta}$. The input of the forward propagation is the dataset $D_i$ and its output is the predicted  parameters of ConvNet $\Omega$  by $H_{\theta}$. Note that  the architecture of $\Omega$ is consistently fixed during both the training and testing stages, e.g., ResNet-18. 
This choice is logical as we often utilize a representative deep model for data from different domains. Therefore, it becomes significantly meaningful to be able to predict the network parameters for unseen image data using an identical network architecture.
$\mathcal{M}^\Omega=\{ \mathcal{M}_i^\Omega \}_{i=1}^{\mathcal{N}}$ denotes the ground-truth parameter set of ConvNet $\Omega$ corresponding to  image datasets  $\mathcal{D}^{train}$, where $\mathcal{M}_i^\Omega$ is the ground-truth parameters for the dataset $D_i$.
$\mathcal{L}$ is a loss function, measuring the difference between the ground-truth parameters $\mathcal{M}_i^\Omega$ and the predicted parameters. 

The core idea in (\ref{obj1}) is to learn a hyper-mapping $H_{\theta}$ between image datasets $\mathcal{D}^{train}$ and the ConvNet parameter set $\mathcal{M}^\Omega$, based on our finding that there are correlations between image datasets and the ConvNet parameters, as shown in Fig. \ref{correlation}.
However, it is prohibitive if preparing thousands of image datasets $D_i$ and training  $\Omega$ on $D_i$ to obtain the corresponding ground-truth parameters  $\mathcal{M}_i^\Omega$ respectively. 
To alleviate this problem, we adopt a meta-learning based~\cite{vinyals2016matching} approach to train the hypernetwork $H_{\theta}$, and propose another objective function as:
\begin{equation}
\label{obj}
\setlength{\abovedisplayskip}{0.1in}
\setlength{\belowdisplayskip}{0.1in}
\mathop{\arg\min}\limits_{\theta} \quad \sum_{i=1}^{\mathcal{N}}\sum_{x_{j} \in D_i} \mathcal{L}(x_j,y_j;\mathcal{F}(D_i,\Omega;H_{\theta})),
\end{equation}
Instead of optimizing $H_{\theta}$ by directly matching the predicted parameters $\mathcal{F}(D_i,\Omega;H_{\theta})$ with the ground-truth parameters $\mathcal{M}_i^\Omega$, we can adopt a typical loss, e.g., cross-entropy, to optimize  $H_{\theta}$, where each dataset $D_i$ can be regarded as a task in meta-learning \cite{vinyals2016matching}.  By learning on multiple tasks,  the parameter predictor $H_{\theta}$ is gradually able to learn to predict performant parameters for training datasets $\mathcal{D}^{train}$.  
During testing, we can utilize $\mathcal{F}(D,\Omega;H_{\theta})$ to directly predict the parameters for a dataset $D$ never seen in $\mathcal{D}^{train}$ with only a single forward propagation without training on $D$.

\subsection{Overview of Our Framework}

Our goal is to learn a hypernetwork $H_{\theta}$ to directly predict the ConvNet parameters for an unseen image dataset. There are two issues that remain unresolved. First, the sizes of different $D_i$ dataset may vary and the dataset sizes themselves can be large,
which poses challenges in training $H_{\theta}$; Second, there may exist correlations among  parameters of different layers within a network. However, how to capture such context relations among parameters has not been fully explored. 

To this end, we propose a novel framework called PudNet, illustrated in Fig. \ref{frame}. PudNet begins with a dataset compression module, which compresses each dataset $D_i$ into a concise sketch $\mathbf{s}_i$, summarizing the key characteristics of $D_i$.
Next, our context-aware parameter prediction module takes the sketch $\mathbf{s}_i$ as input and predicts the parameters of the target network, such as ResNet-18.
Furthermore, multiple adaptive hyper-recurrent units with shared weights are constructed to capture parameter dependencies across different layers of the network. Finally, PudNet is optimized in a meta-learning based manner. 

\subsection{Dataset Compression} 
To address the issue of varying dataset size, we employ a data compression module to compress each image dataset into a fixed-size sketch.
In recent years, many data compression methods have been proposed, including matrix sketching \cite{liberty2013simple,qian2015subset}, random projection \cite{liberty2007randomized}, etc. 
These methods can potentially be applied in our data compression module.
For simplification, we utilize a neural network to extract a feature vector as the representation of each sample. Subsequently, we perform an average pooling operation to generate a final vector as the sketch of the dataset.
The sketch $\mathbf{s}_i$ for dataset $D_i$ can be calculated as: 
\begin{equation}
\setlength{\abovedisplayskip}{0.1in}
\setlength{\belowdisplayskip}{0.1in}
\mathbf{s}_i=\frac{1}{|D_i|}\sum_{x_j \in D_i} T_\phi(x_j), \quad D_i \in \{D_{i}\}_{i=1}^{\mathcal{N}},
\end{equation}
where $T_\phi(\cdot)$ denotes a feature extractor parameterized by $\phi$, and $|D_i|$ is the size of the dataset $D_i$. $\phi$ is jointly trained with PudNet in an end-to-end fashion. 
In our PudNet, the feature extractor contains three basic blocks, each comprising a $5\times5$ convolutional layer, a leakyReLU function, and a batch normalization layer.
For future work, more efforts could be made to explore more effective solutions for summarizing dataset information. One potential direction is to investigate the use of statistic networks \cite{DBLP:conf/iclr/EdwardsS17} to compress datasets.
 
\textbf{Discussion}. In this paper, we adopt a series of widely used image datasets to evaluate the effectiveness of our method. As the number of classes increases, our method may encounter information loss during the dataset compression step, especially when the class distributions are imbalanced. 
In such cases, the dataset sketch may be biased towards the head classes, which have a larger number of samples compared to the other classes.
To alleviate this, we employ a simple clustering-based method. 
Specifically, we first perform clustering on the dataset. Then, during the dataset compression process, we average the cluster centroid features of each class to produce the dataset sketch used for parameter prediction.
By doing so, we can prevent the dataset sketch from being dominated by head classes and mitigate the problem of information loss from the tail classes. 


\subsection{Context-Aware Parameter Prediction} 
After obtaining the sketches for all training datasets, we will feed them into the context-aware parameter prediction module, as shown in Fig. \ref{frame}. In the following, we will introduce this module in detail.


{
\subsubsection{Capturing Contextual Parameter Relations via Adaptive Hyper-Recurrent Units}
First, since we aim to establish a hypernetwork that can capture the hyper-mapping between datasets and their corresponding network parameters, it is essential for the hypernetwork to be adaptable to different datasets. This adaptability allows it to predict dataset-specific parameters effectively.
Second, as the input of a neural network would sequentially  pass forward its layers, 
the parameters of different layers are not independent. 
Neglecting the contextual relations among parameters from different layers may result in sub-optimal solutions.

\begin{figure}
\centering
\includegraphics[width=1\columnwidth]{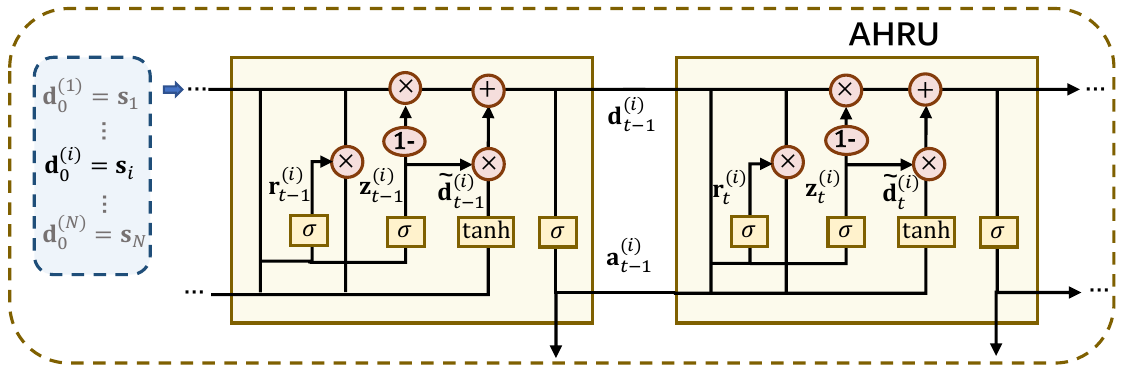}
\caption{An illustration of capturing context relations via AHRU.}
\label{cgru}
\vspace{-0.2in}
\end{figure}

In light of this, we construct Adaptive Hyper-Recurrent Units (AHRU), specifically designed for parameter prediction of ConvNets. AHRU enhances the capabilities of the gated recurrent unit by adapting to different datasets, while also serving as the hyper-recurrent unit to capture context-aware parameter relations during the parameter generation process.
As shown in Fig. \ref{cgru}, we utilize the dataset sketch embeddings $\mathbf{s}_i$ as the input to the hyper-recurrent unit:
\begin{align}
\mathbf{d}_0^{(i)}=\mathbf{s}_i, \quad \mathbf{s}_i \in \{\mathbf{s}_i\}_{i=1}^{\mathcal{N}},
\end{align}
where $\mathbf{d}_0^{(i)}$ represents the initial hidden state of AHRU. In this way, we can integrate dataset-specific information into parameter prediction.

Then, we predict parameter representation $\mathbf{a}_{t}^{(i)}$ of $t$-th layer in $\Omega$ for the $i$-th dataset as follows:
\begin{align}\label{gru} \nonumber
&\mathbf{r}_t^{(i)} = \sigma (\mathbf{W}_r \cdot [\mathbf{d}_{t-1}^{(i)},\mathbf{a}_{t-1}^{(i)}]), \\ \nonumber
&\mathbf{z}_t^{(i)} = \sigma (\mathbf{W}_z \cdot [\mathbf{d}_{t-1}^{(i)},\mathbf{a}_{t-1}^{(i)}]),  \\ \nonumber
&\tilde{\mathbf{d}}_t^{(i)} = tanh(\mathbf{W}_h \cdot [\mathbf{r}_t^{(i)} \ast \mathbf{d}_{t-1}^{(i)},\mathbf{a}_{t-1}^{(i)}]),  \\ \nonumber
&\mathbf{d}_t^{(i)} = (1-\mathbf{z}_t^{(i)})\ast \mathbf{d}_{t-1}^{(i)} + \mathbf{z}_t^{(i)} \ast \tilde{\mathbf{d}}_t^{(i)},   \\ \nonumber
&\mathbf{a}_{t}^{(i)} = \sigma (\mathbf{W}_o \cdot \mathbf{d}_t^{(i)}),  
\end{align}
where  $\mathbf{d}_{t-1}^{(i)}$ serves as the hidden state in AHRU for conveying both dataset-related context information specific to the $i$-th dataset and parameter information from previous layers.
$\mathbf{W}_r, \mathbf{W}_z, \mathbf{W}_h, \mathbf{W}_o$ are learnable parameters shared across different datasets.
In AHRU, the reset gate $\mathbf{r}_t^{(i)}$ decides how much data information in hidden state $\mathbf{d}_{t-1}^{(i)}$ needs to be reset. 
The new memory $\tilde{\mathbf{d}}_t^{(i)}$ absorbs information from both $\mathbf{d}_{t-1}^{(i)}$ and  $\mathbf{a}_{t-1}^{(i)}$. 
The update gate $\mathbf{z}_t$ regulates how much information in $\tilde{\mathbf{d}}_t^{(i)}$ to update and how much information in $\mathbf{d}_{t-1}^{(i)}$ to forget.
During this process, the hidden state $\mathbf{d}_{t-1}^{(i)}$, along with the  parameter representation $\mathbf{a}_{t-1}^{(i)}$ in the previous layer, are utilized to produce $\mathbf{a}_{t}^{(i)}$. By doing so, $\mathbf{a}_{t}^{(i)}$ can effectively encompass the dependency information between different layers while also adapting to the $i$-th dataset. 
The derived parameter representation $\mathbf{a}_{t}^{(i)}$ is then taken as input to the weight generator for predicting the parameters of the $t$-th layer of the target network $\Omega$.

It is worth noting that AHRU is designed based on the GRU-based architecture \cite{DBLP:conf/emnlp/ChoMGBBSB14} but differs from the traditional GRU.
Our AHRU is specifically designed to adapt to various datasets based on dataset sketches, while also capturing dependencies among parameters across different layers.
Specifically, conventional GRU utilizes the randomly initialized hidden state to carry the long-term memory. In each recurrent step, it needs to receive external input (e.g., word) to update the hidden state and capture the contextual information. 
Instead, we employ each dataset sketch embedding as input and follow a self-loop structure to incorporate the predicted structure parameter representation from the previous layer. This approach enables us to leverage dataset-specific information and meanwhile allows the information of shallower layer parameters to assist in predicting parameters in deeper layer. 
By utilizing AHRU, our PudNet possesses the ability to adapt to various datasets while effectively capturing dependencies among parameters across different layers.

\subsubsection{Initial Residual Connection}
To ensure that the final context-aware output retains a portion of the initial dataset information, we present an initial residual connection between the dataset sketch embedding $\mathbf{s}_i$ and $\mathbf{a}_{t}$ as: 
\begin{equation}
\setlength{\abovedisplayskip}{0.1in}
\setlength{\belowdisplayskip}{0.1in}
\hat{\mathbf{a}}_{t}^{(i)} = \mathbf{a}_{t}^{(i)} \times (1-\eta) + \mathbf{s}_i \times \eta,  \quad \hat{\mathbf{a}}_{t}^{(i)} \in \mathbb{R}^{m},
\end{equation}
where $\eta$ is the hyperparameter. After obtaining $\hat{\mathbf{a}}_{t}^{(i)}$, we input $\hat{\mathbf{a}}_{t}^{(i)}$ into the weight generator to generate parameters of the $t$-th layer of the target network $\Omega$.

\subsubsection{Weight Generator}
Since the target network $\Omega$ usually has varying sizes and dimensions across different layers, we construct a weight generator $g_{\psi_t}$ for each layer $t$ to transform the fixed-dimensional $\hat{\mathbf{a}}_{t}^{(i)}$ into a network parameter tensor $\mathbf{w}_t^{(i)}$ with variable dimensions. Here $g_{\psi_t}$ denotes the weight generator for the $t$-th layer, $\psi_t$ is the learnable parameters of $g_{\psi_t}$, and $\mathbf{w}_t^{(i)}$ is the predicted parameter of the $t$-th layer in $\Omega$ for the $i$-th dataset. We can derive the predicted parameter of the $t$-th layer as:
\begin{align}\label{w}
\setlength{\abovedisplayskip}{0.1in}
\setlength{\belowdisplayskip}{0.1in}
\mathbf{w}_{t}^{(i)}=g_{\psi_t}(\hat{\mathbf{a}}_{t}^{(i)}), \quad \mathbf{w}_{t}^{(i)} \in \mathbb{R}^{C_t^{out} \times C_t^{in} \times k_t^2},
\end{align}
where $g_{\psi_t}$ consists of one linear layer and two $1 \times 1$ convolutional layers.
Fig.~\ref{weightgen} shows the architecture of the weight generator. 
The weight generator takes the $m$-dimensional vector $\hat{\mathbf{a}}_{t}^{(i)}$ as input and outputs a tensor with dimensions $C_t^{out} \times C_t^{in} \times k_t^2$. This tensor is the parameters generated for the $t$-th convolutional layer in $\Omega$. In a typical deep convolutional network, the model parameters are in the kernels of convolutional layers. We suppose that the $t$-th convolutional layer contains $C_t^{out} \times C_t^{in}$ kernels and each kernel has dimensions $k_t \times k_t$. 
To generate weights for the convolutional layers, we introduce a weight generator which consists of two parts. Firstly, we take the context-aware dataset embedding as input to generate a weight embedding of size $d=p k_t k_t$. Secondly, the weight embedding is reshaped to $(p, k_t k_t)$ to get $p$ flattened kernels, where $p$ is a hyper-parameter. These $p$ flattened kernels are passed through $1\times1$ convolutional layers and an activation layer to obtain the desired $C_t^{out}C_t^{in}$ flattened kernels. Subsequently, the matrix  with dimensions $C_t^{out}C_t^{in} \times k_t k_t$ is reshaped into a tensor with dimensions $C_t^{out} \times C_t^{in} \times k_t \times k_t$ as the predicted parameter $\mathbf{w}_t^{(i)}$.


\begin{figure}[!h]
\centering
\includegraphics[width=1\columnwidth]{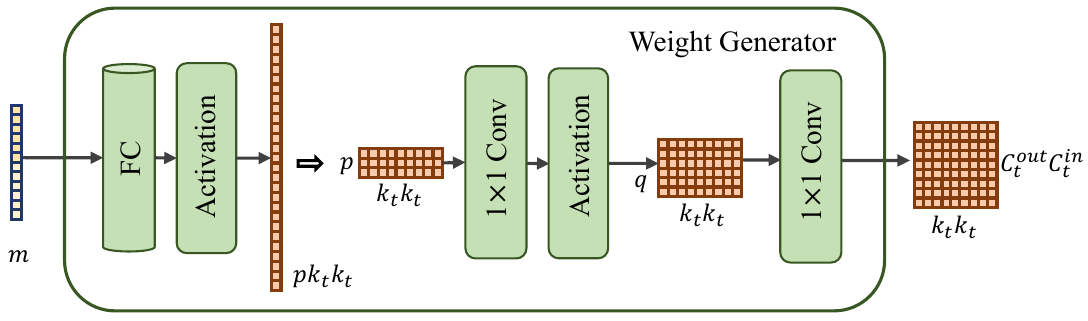}
\vspace{-0.2in}
\caption{Architecture of the weight generator.}
\label{weightgen}
\vspace{-0.1in}
\end{figure}

When the parameters of each layer in the target network are predicted, we can use these predicted parameters  as the final parameters of $\Omega$ for inference. 

\subsection{Optimization of Our Framework}
In this section, we present the optimization process for our PudNet. 
Unlike traditional classification tasks where the training and testing data share identical label spaces, our task involves label spaces that can differ and may not overlap between training and testing datasets.
Thus, training a classification head on the training data cannot be directly used to predict labels of the testing data.
Motivated by several metric learning methods \cite{chen2021meta, oreshkin2018tadam}, we introduce a parameter-free classification method to address the above challenge. 

Similar to \cite{chen2021meta}, we obtain a metric-based category prediction on class $c_k$ as:  
\begin{align}
&p(y=c_k|x_j,\Omega,D_i; H_{\theta}) \nonumber \\ 
&=\frac{exp(\tau \cdot <f(x_j;\mathcal{F}(D_i,\Omega; H_{\theta})),\mathbf{u}_k>)}{\sum_{c} {exp}(\tau \cdot <f(x_j;\mathcal{F}(D_i,\Omega; H_{\theta})),\mathbf{u}_c>)},  
\label{metric_eq}
\end{align}
where the centroid of class $c_k$, denoted by $\mathbf{u}_k$, represents the average output of the predicted network $\Omega$ over samples belonging to class $c_k$, as described in \cite{snell2017prototypical}.
$<\cdot,\cdot>$ denotes the cosine similarity between two vectors, and $\tau$ is a learnable temperature parameter. $f(x_j;\mathcal{F}(D_i,\Omega; H_{\theta}))$ is the output of the target network $\Omega$ based on the input $x_j$.
Then, the parameter-free classification loss can be defined as follows: 
\begin{align}
\mathcal{L}_{1}=\sum_{i=1}^{\mathcal{N}}\sum_{x_j \in D_i} \mathcal{L}(p(y|x_j,\Omega,D_i;H_{\theta}), y_j),
\end{align}
where $y_j$ is the true label of $x_j$, $\mathcal{L}$ is the cross-entropy loss. 

To further improve the performance of the model, we incorporate a full classification head $\mathcal{Q}_{\varphi}$, parameterized by $\varphi$, as an auxiliary task during the training of our hypernetwork, motivated by \cite{oreshkin2018tadam}. 
The full classification head aims to map the output of the target network $\Omega$ into  probabilities of whole classes $\mathcal{C}^{tr}$ in the dataset $\mathcal{D}^{train}$. 
The full classification loss is defined as:
\begin{align}
\mathcal{L}_{2}=\sum_{i=1}^{\mathcal{N}}\sum_{x_j \in D_i} \mathcal{L}(\mathcal{Q_{\varphi}}(f(x_j;\mathcal{F}(D_i,\Omega; H_{\theta}))), y_j).
\end{align}


Furthermore, to ensure the consistency between parameter-free classification prediction and full classification prediction, as motivated by \cite{chen2022asm2tv,wu2019large}, we introduce a Kullback-Leibler Divergence loss to encourage the alignment of their predicted probabilities:
\begin{align}
\mathcal{L}_{3}=\sum_{i=1}^{\mathcal{N}}\sum_{x_j\in D_i} KL(q(y|x_j)||p(y|x_j)),
\end{align}
where $KL$ is the Kullback-Leibler Divergence. $p(y|x_j)$ and $q(y|x_j)$ denote the predicted probabilities of $x_j$ obtained from the parameter-free based and full classification based methods, respectively. 
To ensure dimensional compatibility, the probabilities of the corresponding classes in $p(y|x_j)$ are padded with zeros to match the dimension of $q(y|x_j)$.

Finally, we give the overall multi-task loss as:
\begin{equation}
\label{final}
\setlength{\abovedisplayskip}{0.1in}
\setlength{\belowdisplayskip}{0.1in}
\mathcal{L}_{train}=\mathcal{L}_{1}+\mathcal{L}_{2}+\mathcal{L}_{3}.
\end{equation}
where $D_i$ can be regarded as a task analogous to it in meta-learning. By minimizing (\ref{final}), our hypernetwork can be effectively trained. 
During testing, for unseen data, we employ our hypernetwork to directly predict its network parameters and utilize the parameter-free classification method for classification. 

We provide the training procedure of our PudNet in Algorithm 1. For each training dataset $D_i \in \mathcal{D}^{train}$, we first derive the sketch $\mathbf{s}_i$ of dataset $D_i$ and initialize the hidden state as $\mathbf{d}_0^{(i)}=\mathbf{s}_i$ in AHRU. Then, we predict the parameters of each layer in the target network $\Omega$. Finally, we optimize the learnable parameters $\theta$ and $\varphi$ using the overall multi-task loss $\mathcal{L}_{train}$. We also present the inference process of our PudNet for an unseen dataset in Algorithm 2.

\begin{algorithm}
\label{algori}
\caption{The Training of PudNet}
\KwIn{A set of training datasets $\mathcal{D}^{train}=\{ D_i \}_{i=1}^{\mathcal{N}}$, target network architecture 
 $\Omega$. }
\ \ Initialize the learnable parameter $\theta$ of $H_{\theta}$.\\ 
\ \ Initialize the learnable parameter $\varphi$ of the auxiliary full classification.\\
\While{not converged}{
 \For{ $i \in \{1,\cdots,\mathcal{N}\}$ }{
    \ \ Obtain a sketch $\mathbf{s}_i$ of dataset $D_i$ via the dataset compression module; \\
    \ \ Initialize the hidden state $\mathbf{d}_0^{(i)}=\mathbf{s}_i$; \\
    \For{ each layer $t$ in $\Omega$}{
         Predict the parameters $w_t^{(i)}$ of the $t$-th layer by weight generator;\\
         Fix the predicted parameters $w_t^{(i)}$ to $\Omega$;
    }
    \For{ each batch $b$ in $D_i$}{
         Compute the parameter-free loss $\mathcal{L}_1$ with a  batch size of $b$;\\
         Compute the full classification loss $\mathcal{L}_2$ and consistency loss $\mathcal{L}_3$ with a  batch size of $b$;\\
         Update the learnable parameters $\theta$, $\varphi$ by the overall multi-task loss $\mathcal{L}_{train}$ ;
        
    }
 }
}
\KwOut{The PudNet $H_\theta$.}
\end{algorithm}

\begin{algorithm}
\label{algori}
\caption{The Inference of PudNet for Unseen Data}
\KwIn{An unseen dataset $D_{i} \in \mathcal{D}^{test}$, the learned PudNet $H_{\theta}$, target network architecture $\Omega$. }
\For{ each layer $t$ in $\Omega$}{
    \ \ Predict the parameters $w_t^{(i)}$ of the $t$-th layer by the PudNet $H_\theta$;\\
    \ \ Fix the predicted parameters $w_t^{(i)}$ to $\Omega$;
}
    \ \ Conduct inference for dataset $D_{i}$ with parameterized network $\Omega$ ;\\  
\KwOut{The prediction results for dataset $D_{i}$.}
\end{algorithm}

\subsection{Theoretical Analysis of the Hyper-Mapping }
We provide a theoretical analysis elucidating how the model learns the hyper-mapping between datasets and their corresponding network parameters, which offers interpretability. 

Recall that for each image dataset $D_i \in \mathcal{D}^{train}$, the learning of the hyper-mapping  between dataset $D_i$ and their corresponding network parameters $\mathcal{M}_i$ can be formulated as: leveraging a hyper-network $H_{\theta}$ to produce an optimal model $\Omega_{\mathcal{M}_i^*}$ with  parameters $\mathcal{M}_i^*=H_{\theta}(D_i)$ via optimizing $\mathop{\arg\min}\limits_{\theta}\sum_{D_i \in \mathcal{D}^{train}}\mathcal{L}(\Omega_{\mathcal{M}_i^*},D_i)$. To write conveniently, we denote $D_i$ as $d$.  We can learn the hypernetwork $H_{\theta}$  by solving an optimization problem:  
$$\mathop{\arg\min}\limits_{\theta}\mathbb{E}_{d \sim p(d)}\mathcal{L}(H_{\theta}(d),d).$$

Solving $H_{\theta}$ is generally feasible. We denote $H_{\theta}$ as $h$ and optimize the problem in the form $h^{*} :=\mathop{\arg\min}_{h}\mathcal{L}(h(d),d)$. 
For $h$, conditioning $\mathcal{L}(h^*(d),d)$ is an extremum for any infinitesimal $\delta d$, we can deduce $h^*$  using the ordinary differential equation:
$$
\frac{\partial \mathcal{L}}{\partial {h}} (h^*(d+\delta d),d+\delta d)=\frac{\partial ^{2} \mathcal{L}}{\partial h^2} \frac{\partial h^*}{\partial d} \delta d + \frac{\partial ^{2} \mathcal{L}}{\partial h \partial d}  \delta d + O(\delta d)=0.
$$
Then we can get:
\begin{align}
\frac{\partial ^{2} \mathcal{L}}{\partial h^2} \frac{\partial h^*}{\partial d} = - \frac{\partial ^{2} \mathcal{L}}{\partial h \partial d}. 
\end{align}

We proceed by employing the product rule and chain rule to calculate the derivative $\frac{\partial h^*}{\partial d}$. We substitute $\frac{\partial \mathcal{L}}{\partial h}=\frac{\partial \mathcal{L}}{\partial y}\frac{\partial y}{\partial h}$ into the equation, where $y$ represents the output of the ConvNet model. 
Given these, $\frac{\partial^2 \mathcal{L}}{\partial h^2}$ is the Hessian matrix of $\mathcal{L}$ with respect to $h$ as:
\begin{align}
\frac{\partial ^{2} \mathcal{L}}{\partial {{h^2}}}=\frac{\partial ^{2} \mathcal{L}}{\partial y^2}(\frac{\partial y}{\partial h})^2+\frac{\partial \mathcal{L}}{\partial {y}}\frac{\partial ^2y}{\partial {h^2}},    
\end{align}
and we have:
\begin{align}
\frac{\partial ^{2} \mathcal{L}}{\partial {{h}} \partial {{d}}} = &(\frac{\partial ^{2} \mathcal{L}}{\partial y^2}\frac{\partial y}{\partial d}+\frac{\partial ^{2} \mathcal{L}}{\partial y^2}\frac{\partial y}{\partial h}\frac{\partial h}{\partial d})\frac{\partial y}{\partial h} \nonumber \\
&+ \frac{\partial \mathcal{L}}{\partial y}(\frac{\partial ^2 y}{\partial h \partial d}+\frac{\partial ^2y}{\partial {h^2}}\frac{\partial h}{\partial d})
\end{align}

{
Assuming that the Hessian of $\mathcal{L}$ is non-singular, thus we can solve for 
$\frac{\partial h^*}{\partial d}$ as: 
$$\frac{\partial h^*}{\partial d}=-\frac{1}{2}(\frac{\partial ^{2} \mathcal{L}}{\partial {{y^2}}}(\frac{\partial {y}}{\partial {h}})^2+\frac{\partial \mathcal{L}}{\partial {y}}\frac{\partial ^2y}{\partial {h^2}})^{-1}(\frac{\partial ^2 \mathcal{L}}{\partial {y^2}}\frac{\partial y}{\partial d}\frac{\partial y}{\partial h}+\frac{\partial \mathcal{L}}{\partial y}\frac{\partial ^2 y}{\partial h \partial d}),$$
where the superscript ${-1}$ denotes the matrix inversion.}

Suppose we have found a solution $h^*$ for the optimization problem $\mathop{\arg\min}_{h^*} \mathcal{L}(h^*,d_0)$ concerning the dataset $d=d_0$. We intend to utilize this solution to "track" the local minimum of the loss $\mathcal{L}$ as we move from $d_0$ to another dataset $d=d_1$ within a small vicinity of $d_0$, where the loss $\mathcal{L}$ remains convex and the Hessian of the loss $\mathcal{L}$ is non-singular. 

To accomplish this, we can employ an ordinary differential equation (ODE) approach to integrate the solution along a path $\hat{d}$ that connects the two datasets $d_0$ and $d_1$. 
To track the local minimum of the loss function when transitioning from $d_0$ to $d_1$ along the path $\hat{d}$, we need to numerically integrate the above path $\hat{d}$ with $\partial h^*/\partial d$ as the derivative, where $d_0 = \hat{d}(0)$ and $d_1 = \hat{d}(1)$. By solving this ODE numerically, we obtain the solution corresponding to the local minimum of the loss $\mathcal{L}$:
\begin{align}
\frac{d {h^{*}}}{d \gamma} &= \frac{\partial h^*}{\partial \hat{d}}\frac{d \hat{d}}{d \gamma}  \nonumber \\ 
=& -\frac{1}{2}(\frac{\partial ^{2} \mathcal{L}}{\partial {{y^2}}}(\frac{\partial {y}}{\partial {h}})^2+\frac{\partial \mathcal{L}}{\partial {y}}\frac{\partial ^2y}{\partial {h^2}})^{-1}\frac{\partial ^2 \mathcal{L}}{\partial {y^2}}\frac{\partial y}{\partial \hat{d}}\frac{\partial y}{\partial h} \frac{d \hat{d}}{d \gamma} \nonumber \\ 
&-\frac{1}{2}(\frac{\partial ^{2} \mathcal{L}}{\partial {{y^2}}}(\frac{\partial {y}}{\partial {h}})^2+\frac{\partial \mathcal{L}}{\partial {y}}\frac{\partial ^2y}{\partial {h^2}})^{-1}\frac{\partial \mathcal{L}}{\partial y}\frac{\partial ^2 y}{\partial h \partial \hat{d}}\frac{d \hat{d}}{d \gamma},
\end{align}
where $\gamma$ changes from 0 to 1, and all derivatives are computed at $\hat{d}(\gamma)$ and $h^*(\gamma)$.

The aforementioned equation provides us the way to obtain the empirical solution of $H_{\theta}$ in relation to the proposed PudNet. These concepts establish a theoretical basis and enhance the interpretability of hyper-mapping learning between datasets and their corresponding network parameters.

\section{Experiment}
\subsection{Dataset Construction} 
In the experiment, we evaluate our method by constructing numerous datasets based on several image datasets: {Fashion-MNIST}~\cite{xiao2017fashion}, {CIFAR-100}~\cite{krizhevsky2009learning}, {Mini-ImageNet}~\cite{vinyals2016matching}, {Animals-10}~\cite{gupta2022adjusting}, {CIFAR-10}~\cite{krizhevsky2009learning}, {DTD}~\cite{cimpoi2014describing}, {and a large-scale dataset, ImageNet-1K ~\cite{deng2009ImageNet}}.
These constructed datasets are based on two distinct settings: the \emph{Intra-dataset} prediction setting and the \emph{Inter-dataset} prediction setting.

The \textbf{intra-dataset prediction setting} is described below: 


\textbf{Fashion-set}: 
We construct 2000 groups of datasets from the samples of 6 classes in Fashion-MNIST to train PudNet.
The testing groups of datasets are constructed from samples of the remaining 4 classes. We construct 500 groups of datasets from the 4-category testing set to generate 500 groups of network parameters. 
For each group of network parameters, we also construct another dataset having identical labels but not having overlapped samples with the dataset used for generating parameters, in order to test the performance of the predicted network parameters. 

\textbf{CIFAR-100-set}: 
We select 80 classes from CIFAR-100 as the training sets and the remaining 20 non-overlapping classes as the testing sets. We randomly sample 100,000 groups of datasets for training. Similar to Fashion-set, we construct 500 groups of datasets for parameter prediction, and another 500 groups for testing.

\textbf{ImageNet-100-set}: Similar to CIFAR-100-set, we randomly sample 50,000 groups of datasets from 80-category training sets of Mini-ImageNet to train PudNet. 
We construct 500 groups of datasets from the remaining 20-category testing sets to predict  parameters, and we sample another 500 groups of datasets for testing.

\textbf{ImageNet-1K-set}: We partition ImageNet-1K into two mutually exclusive sets of 800 classes for training and 200 classes for testing. Within the training set of 800 classes, we construct 20,000 sub-datasets to train PudNet. We employ the testing data of 200 classes to evaluate the performance of PudNet. 

\emph{Please note that the training class set and testing class set in the aforementioned datasets are not overlapped.} 


To further validate the performance of our PudNet, we introduce an \textbf{inter-dataset setting} by constructing cross-domain datasets. We construct three training datasets: CIFAR-100-set with 100,000 dataset groups, ImageNet-100-set with 50,000 dataset groups, and ImageNet-1K-set with 20,000 dataset groups.  Each of these datasets is used individually to train PudNet.
Samples from Animals-10, CIFAR-10, and DTD are utilized for evaluation, respectively.
Note that ImageNet and the above three datasets are popular cross-domain datasets, which have been widely utilized for out-of-distribution (OOD) learning \cite{huang2021trash,bibas2021single} and cross-domain learning \cite{li2022cross,islam2021broad}.
The summary of the constructed datasets is as follows:

\textbf{CIFAR-100$\to$Animals-10, ImageNet-100$\to$Animals-10, ImageNet-1k$\to$Animals-10}: We randomly select 80\% of the samples from Animals-10 and input them into PudNet, which is trained on CIFAR-100-set, ImageNet-100-set, and ImageNet-1K-set, respectively, to generate network parameters. The remaining 20\% of samples are then used to test the performance of the predicted network parameters.

\textbf{CIFAR-100$\to$CIFAR-10, ImageNet-100$\to$CIFAR-10, ImageNet-1k$\to$CIFAR-10}: We randomly select 50,000 samples from CIFAR-10 and input them into PudNet, which is trained on CIFAR-100-set, ImageNet-100-set, and ImageNet-1K-set, respectively, to generate parameters. The remaining 10,000 samples are utilized for testing. 

\textbf{CIFAR-100$\to$DTD, ImageNet-100$\to$DTD, ImageNet-1k$\to$DTD}}: DTD is an image texture classification dataset \cite{cimpoi2014describing}, which is included in the visual adaptation benchmark \cite{zhai2019large}. The labels in DTD differ significantly from those in ImageNet. We randomly choose 2/3 of the samples from DTD to generate parameters, while the rest 1/3 is for testing.




\begin{table*}[t]
\setlength\tabcolsep{5pt}
\renewcommand\arraystretch{1} 
\small
\centering
\caption{Results of different methods in terms of the target network ResNet-18 in the intra-dataset setting.}
\begin{tabular}{@{}clcccccc@{}}
\toprule
\multicolumn{2}{c}{Method}                             & \multicolumn{2}{c}{Fashion-set}       & \multicolumn{2}{c}{CIFAR-100-set}        & \multicolumn{2}{c}{ImageNet-100-set}  \\ 
\multicolumn{2}{c}{}                                   & ACC(\%)            & time (sec.)   & ACC            & time (sec.)   & ACC            & time (sec.)   \\ \midrule
\multicolumn{2}{c}{Pretrained}    & 93.76$\pm$0.47          & -        & 64.58$\pm$0.59          & -     & 65.67$\pm$0.73          & -        \\  \midrule
\multicolumn{2}{c}{MatchNet \cite{vinyals2016matching}}      & 90.16$\pm$0.53          & -        & 56.23$\pm$0.71          & -     & 53.17$\pm$0.91          & -        \\
\multicolumn{2}{c}{ProtoNet \cite{snell2017prototypical}}      & 93.64$\pm$0.47          & -        & 60.29$\pm$0.59          & -     & 58.95$\pm$0.83          & -        \\
\multicolumn{2}{c}{Meta-Baseline \cite{chen2021meta}} & 95.35$\pm$0.29 & -        & 67.51$\pm$0.55          & -     & 67.16$\pm$0.70          & -        \\
\multicolumn{2}{c}{Meta-DeepDBC \cite{xie2022joint}}  & 94.28$\pm$0.31          & -        & 69.54$\pm$0.49 & -     & 68.48$\pm$0.60 & -        \\ 
\multicolumn{2}{c}{MUSML \cite{DBLP:conf/icml/JiangK022}}  & 95.87$\pm$0.44          & 2.55        & 66.47$\pm$0.63          & 2.59     & 66.03$\pm$0.91     & 2.60        \\ \midrule

\multirow{3}{*}{\begin{tabular}[c]{@{}c@{}}Adam\\ Scratch\\\cite{DBLP:journals/corr/KingmaB14}\end{tabular}}&1 epoch   & 93.98$\pm$1.21         & 1.83     & 52.82$\pm$1.01    & 3.96       & 46.43$\pm$1.18 & 4.81     \\
                                     &30 epochs                    & 99.91$\pm$0.05         & 54.22    & 74.56$\pm$0.45    & 118.87     & 71.84$\pm$0.69 & 140.37   \\
                                     &50 epochs                    & 99.87$\pm$0.11         & 91.19    & 79.85$\pm$0.47    & 198.12     & 75.98$\pm$0.71 & 231.63   \\ \midrule
\multirow{3}{*}{\begin{tabular}[c]{@{}c@{}}GC \\Scratch\\ \cite{yong2020gradient} \end{tabular}} &1 epoch  & 94.11$\pm$1.25    & 1.88    & 53.21$\pm$1.23  & 4.01 & 47.55$\pm$1.33          & 4.82     \\
&30 epochs                    & 99.94$\pm$0.05         & 54.93    & 75.74$\pm$0.59 & 119.03    & 72.89$\pm$0.73   & 140.98    \\
&50 epochs                    & 99.96$\pm$0.03         & 91.73    & 79.98$\pm$0.55 & 199.61    & 76.73$\pm$0.87   & 232.57   \\ \midrule
\multicolumn{2}{c}{PudNet}  & \textbf{96.24}$\pm$0.39 & 0.50     & \textbf{73.33}$\pm$0.54 & 0.49      & \textbf{71.57}$\pm$0.71 & 0.50  \\ \bottomrule  
\end{tabular}
\label{intra-data}
\vspace{-0.1in}
\end{table*}

\begin{table*}[t]
\small
\renewcommand\arraystretch{1}
\centering
\caption{Results of different methods in terms of the target network ResNet-18 in the inter-dataset setting.}
\begin{tabular}{@{}ccccccccc@{}}
\toprule
\multicolumn{2}{c}{Method}             & \multicolumn{2}{c}{CIFAR-100$\to$Animals-10} & \multicolumn{2}{c}{CIFAR-100$\to$CIFAR-10}   & \multicolumn{2}{c}{CIFAR-100$\to$DTD}                \\
\multicolumn{2}{c}{}                   & ACC (\%)     & time (sec.)  & ACC (\%)     & time (sec.) & ACC (\%)     & time (sec.)\\ \midrule
\multicolumn{2}{c}{Pretrained}         & 33.36$\pm$0.75         & -         & 40.93$\pm$0.48   & -  & 30.27±0.58   & -\\ \midrule
\multicolumn{2}{c}{MatchNet \cite{vinyals2016matching}}  & 33.08$\pm$0.70         & -         & 39.82$\pm$0.69   & -  & 27.17$\pm$0.73   & -\\
\multicolumn{2}{c}{ProtoNet \cite{snell2017prototypical}}& 36.33$\pm$0.60       & -         & 43.22$\pm$0.54   & -  & 32.66$\pm$0.63   & -\\  
\multicolumn{2}{c}{Meta-Baseline \cite{chen2021meta}}    & 38.27$\pm$0.53        & -         & 45.43$\pm$0.59   & -  & 34.25$\pm$0.58   & -\\  
\multicolumn{2}{c}{Meta-DeepDBC \cite{xie2022joint}}     & 40.50$\pm$0.64   & -         & 47.15$\pm$0.63   & -  & 35.50$\pm$0.71   & -\\
\multicolumn{2}{c}{MUSML \cite{DBLP:conf/icml/JiangK022}}& 37.78$\pm$0.59   & 3.21      & 45.37$\pm$0.57  & 5.34 & 33.34±0.68   & 2.18 \\ \midrule
\multirow{4}{*}{\begin{tabular}[c]{@{}c@{}}Adam\\Scratch\\\cite{DBLP:journals/corr/KingmaB14}\end{tabular}} 
&1 epoch    & 22.09$\pm$0.58   & 31.29     & 21.37$\pm$0.71   & 60.79  & 18.83$\pm$0.19   & 4.03 \\
&5 epochs    & 49.12$\pm$0.08   & 156.56    & 33.76$\pm$0.44   & 310.29  & 24.17$\pm$0.72   & 26.01 \\
&10 epochs   & 66.44$\pm$0.37   & 311.74    & 48.21$\pm$0.25   & 621.75  & 33.26$\pm$0.90   & 51.79 \\
&20 epochs   & 73.47$\pm$0.67       & 623.92    & 65.17$\pm$0.55   & 1248.23  & 42.51$\pm$0.44   & 102.11 \\   
&50 epochs   & 77.81$\pm$0.34       & 1558.51   & 80.25$\pm$0.61  & 3109.87  & 52.16$\pm$0.56   & 257.91 \\ \midrule 
\multirow{4}{*}{\begin{tabular}[c]{@{}c@{}}GC\\Scratch \\ \cite{yong2020gradient}\end{tabular}} 
&1 epoch     & 23.01$\pm$1.02 & 30.79 & 21.44$\pm$1.02 & 60.99 & 18.67$\pm$0.34 & 4.01 \\
&5 epochs     & 49.77$\pm$0.54 & 155.34 & 34.41$\pm$0.78 & 310.33 & 24.43$\pm$0.83 & 25.87 \\
&10 epochs    & 68.56$\pm$0.39 & 310.29 & 49.89$\pm$0.53 & 622.01 & 35.49$\pm$0.61 & 51.70 \\
&20 epochs    & 75.04$\pm$0.61 & 623.33  & 66.78$\pm$0.47 & 1248.98 & 44.05±0.52 & 102.63 \\  
&50 epochs    & 77.98$\pm$0.45 & 1557.76 & 82.03$\pm$0.42 & 3109.14 & 53.58±0.61 & 257.24 \\ \midrule 
\multicolumn{2}{c}{PudNet}  & \textbf{43.21}$\pm$0.69 & 0.49 & \textbf{51.05}$\pm$0.56 & 0.56 & \textbf{38.05}$\pm$0.73 & 0.39 \\ \bottomrule
\end{tabular} 
\label{cifar-inter-res18}
\vspace{-0.1in}
\end{table*}

\subsection{Baselines and Implementation Details} We compare our method with the most related work, i.e., traditional iterative based training paradigm such as training from scratch with Adam~\cite{DBLP:journals/corr/KingmaB14} and a training acceleration method, GC \cite{yong2020gradient}. We also consider the pretrained model as a baseline for comparison.
Furthermore, we compare our method with various meta-learning methods, namely MatchNet~\cite{vinyals2016matching}, ProtoNet~\cite{snell2017prototypical}, Meta-Baseline~\cite{chen2021meta}, Meta-DeepDBC~\cite{xie2022joint}, and MUSML~\cite{DBLP:conf/icml/JiangK022}.
For our target network $\Omega$, We explore three different architectures: a 3-layer CNN (ConvNet-3), ResNet-18\cite{he2016deep}, and ResNet-34\cite{he2016deep}. 
Throughout the experiments, we utilize the accuracy (ACC) metric to evaluate the classification performance. Top-1 Accuracy is employed as the evaluation metric, unless otherwise specified.



We perform the experiments using GeForce RTX 3090 Ti GPU. 
All experiments are optimized by the Adam optimizer. We set the learning rate to 0.001 and train PudNet until convergence. In the metric-based learning process, following \cite{chen2021meta}, the temperature $\tau=10$ in Eq.(\ref{metric_eq}) is fixed. 10 labeled samples per class are used to deduce the class centroid.  
As mentioned before, we introduce auxiliary tasks to assist optimization. We incorporate a full classification linear layer (e.g. an 80-way linear head in CIFAR-100) to maintain a static class set during training. 
We also introduce a consistency loss to address the mismatch in dimensions between the logits derived from metric-based classification (e.g., 5-dimensional in CIFAR-100) and the logits produced by the full linear head (e.g., 80-dimensional in CIFAR-100). To align the dimensions, we transpose the 5-dimensional logits to the 80-dimensional logits by padding the remaining values with zeros.
We search $\eta$ from $\{0,0.1,0.2,0.3,\cdots,0.9\}$. 
For the ResNet-18 target network, we set $\eta=0.2$ for Fashion-set, $\eta=0.5$ for CIFAR-100-set, ImageNet-100-set and ImageNet-1K-set. 
The hyperparameter $\eta$ is set as 0.5 in the inter-dataset setting. 
For the ConvNet-3 target network, we set $\eta=0.1$, and for the ResNet-34 target network, we set $\eta=0.5$.

\begin{table*}[t]
\small
\renewcommand\arraystretch{1}
\centering
\caption{Results of different methods in terms of the target network ResNet-18 in the inter-dataset setting.}
\begin{tabular}{@{}ccccccccc@{}}
\toprule
\multicolumn{2}{c}{Method}             & \multicolumn{2}{c}{ImageNet-100$\to$Animals-10} & \multicolumn{2}{c}{ImageNet-100$\to$CIFAR-10}   & \multicolumn{2}{c}{ImageNet-100$\to$DTD}                \\
\multicolumn{2}{c}{}                   & ACC (\%)     & time (sec.)  & ACC (\%)     & time (sec.) & ACC (\%)     & time (sec.)\\ \midrule
\multicolumn{2}{c}{Pretrained}         & 34.79$\pm$0.49         & -         & 34.54$\pm$0.63   & -  & 31.05$\pm$0.51   & -\\ \midrule
\multicolumn{2}{c}{MatchNet \cite{vinyals2016matching}}         & 32.38$\pm$0.82         & -         & 31.98$\pm$0.72   & -  & 28.15$\pm$0.67   & -\\
\multicolumn{2}{c}{ProtoNet \cite{snell2017prototypical}}         & 35.78$\pm$0.64        & -         & 35.21$\pm$0.74   & -  & 31.58$\pm$0.77   & -\\  
\multicolumn{2}{c}{Meta-Baseline \cite{chen2021meta}}         & 36.21$\pm$0.42        & -         & 38.34$\pm$0.69   & -  & 33.07$\pm$0.83   & -\\  
\multicolumn{2}{c}{Meta-DeepDBC \cite{xie2022joint}}       & 38.57$\pm$0.55         & -         & 40.93$\pm$0.71   & -  & 34.01$\pm$0.69   & -\\
\multicolumn{2}{c}{MUSML \cite{DBLP:conf/icml/JiangK022}}              & 36.89$\pm$0.44         & 3.21      & 37.12$\pm$0.65  & 5.34 & 33.67$\pm$0.63   & 2.18 \\ \midrule
\multirow{4}{*}{\begin{tabular}[c]{@{}c@{}}Adam\\Scratch\\\cite{DBLP:journals/corr/KingmaB14}\end{tabular}} &1 epoch     & 22.09$\pm$0.58         & 31.29     & 21.37$\pm$0.71   & 60.79  & 18.83$\pm$0.19   & 4.03\\
&5 epochs    & 49.12$\pm$0.08         & 156.56    & 33.76$\pm$0.44   & 310.79  & 24.17$\pm$0.72   & 26.01\\
&10 epochs   & 66.44$\pm$0.37         & 311.74    & 48.21$\pm$0.25   & 621.75  & 33.26$\pm$0.90   & 51.79\\
&20 epochs   & 73.47$\pm$0.67         & 623.92    & 65.17$\pm$0.55   & 1248.23  & 42.51$\pm$0.44   & 102.11\\   
&50 epochs   & 77.81$\pm$0.34         & 1558.51   & 80.25$\pm$0.61   & 3109.87  & 52.16$\pm$0.56   & 257.91\\ \midrule 
\multirow{4}{*}{\begin{tabular}[c]{@{}c@{}}GC\\Scratch \\ \cite{yong2020gradient}\end{tabular}} &1 epoch     & 23.01$\pm$1.02          & 30.79   & 21.44$\pm$1.02         & 60.99     & 18.67$\pm$0.34   & 4.01\\
&5 epochs     & 49.77$\pm$0.54 & 155.34  & 34.41$\pm$0.78    & 310.33   & 24.43$\pm$0.83   & 25.87  \\
&10 epochs    & 68.56$\pm$0.39 & 310.29  & 49.89$\pm$0.53    & 622.01   & 35.49$\pm$0.61   & 51.70 \\
&20 epochs    & 75.04$\pm$0.61 & 623.33 & 66.78$\pm$0.47    & 1248.98   & 44.05$\pm$0.52   & 102.63\\  
&50 epochs   & 77.98$\pm$0.45  & 1557.76    & 82.03$\pm$0.42   & 3109.14  & 53.58$\pm$0.61   & 257.24\\ \midrule 
\multicolumn{2}{c}{PudNet}               & \textbf{42.43}$\pm$0.58 & 0.48     & \textbf{45.07}$\pm$0.70   & 0.57 & \textbf{47.50}$\pm$0.71   & 0.39\\ \bottomrule
\end{tabular} 
\label{mini-cross}
\vspace{-0.1in}
\end{table*}

\begin{table}[h]
\small
\setlength\tabcolsep{6pt}
\renewcommand\arraystretch{1}
\centering
\caption{Results of different methods on ImageNet-1K-set in the intra-dataset setting.}
\begin{tabular}{@{}cccc@{}}
\toprule
\multicolumn{2}{c}{Method}        & \multicolumn{2}{c}{ImageNet-1K-set} \\ 
\multicolumn{2}{c}{}              & ACC(\%)       & time(sec.)      \\ \midrule
\multicolumn{2}{c}{Pretrained}    & 35.77$\pm$0.84   & -               \\ \midrule
\multicolumn{2}{c}{Meta-Baseline \cite{chen2021meta}} & 38.92$\pm$0.93   & -               \\
\multicolumn{2}{c}{Meta-DeepDBC  \cite{xie2022joint}}  & 39.15$\pm$1.12   & -               \\
\multicolumn{2}{c}{MH \cite{zhao2020meta}}         & 39.07$\pm$0.89   & 39.22            \\ 
\multicolumn{2}{c}{MUSML \cite{DBLP:conf/icml/JiangK022}}         & 41.33$\pm$0.95   & 35.17            \\ 
\multicolumn{2}{c}{GAP \cite{kang2023meta}}         & 42.57$\pm$0.87   &  36.21           \\\midrule
                & 100 steps        & 9.35 $\pm$1.42   & 229.33           \\
GC              & 2000 steps       & 37.48$\pm$0.81   & 4488.71         \\
Scratch         & 4000 steps      & 44.65$\pm$0.97   & 8967.55         \\
\cite{yong2020gradient}& 6000 steps       & 52.03$\pm$0.85   & 13453.29        \\ \midrule
\multicolumn{2}{c}{PudNet}        & \textbf{44.92}$\pm$0.91   & 3.89            \\ \bottomrule
\end{tabular}
\label{IN-1K-intra}
\vspace{-0.2in}
\end{table}

\subsection{Result and Analysis}
\textbf{Performance Analysis of Intra-dataset Prediction.} 
Table~\ref{intra-data} shows the results of our method in the intra-dataset setting.
We could find that our method consistently outperforms both the  meta-learning methods and the pretrained method. 
This demonstrates the effectiveness of learning a hyper-mapping between datasets and their corresponding network parameters.
To showcase the time savings achieved by our method, we provide the training time required for training the model from scratch using the widely-used optimizer, Adam, as well as the training acceleration technique, GC. 
We observe that it takes around 55, 119, and 141 GPU seconds to train ResNet-18 using the accelerated method GC and the network obtains top-1 accuracies of 99.94\%, 75.74\%, and 72.89\%  on Fashion-set, CIFAR-100-set, and ImageNet-100-set respectively. 
In contrast, our method only costs around 0.5 GPU seconds to predict the parameters of ResNet-18, while still achieving comparable performance (96.24\%, 73.33\%, and 71.57\% top-1 accuracies) on the three datasets respectively, at least 100 times faster than the accelerated method.

\begin{table*}[!h]
\small
\renewcommand\arraystretch{1}
\centering
\caption{Results of different methods on ImageNet-1K in the inter-dataset setting.}
\begin{tabular}{@{}ccccccccc@{}}
\toprule
\multicolumn{2}{c}{Method}             & \multicolumn{2}{c}{ImageNet-1K$\to$Animals-10} & \multicolumn{2}{c}{ImageNet-1K$\to$CIFAR-10}   & \multicolumn{2}{c}{ImageNet-1K$\to$DTD}                \\
\multicolumn{2}{c}{}                   & ACC (\%)     & time (sec.)  & ACC (\%)     & time (sec.) & ACC (\%)     & time (sec.)\\ \midrule
\multicolumn{2}{c}{Pretrained}         & 51.02$\pm$0.54         & -         & 40.31$\pm$0.69   & -  & 44.19$\pm$0.61   & -\\ \midrule
\multicolumn{2}{c}{Meta-Baseline \cite{chen2021meta}}      & 52.19$\pm$0.63   & -         & 42.66$\pm$0.72   & -  & 45.43$\pm$0.89   & -\\ 
\multicolumn{2}{c}{Meta-DeepDBC \cite{xie2022joint}}       & 54.37$\pm$0.59       & -         & 44.74$\pm$0.70   & -  & 46.53$\pm$0.72   & -\\
\multicolumn{2}{c}{MH \cite{zhao2020meta}}  & 53.14$\pm$0.64   &  5.35     & 42.97$\pm$0.66  & 7.75 & 45.48$\pm$0.73   & 4.52  \\
\multicolumn{2}{c}{MUSML \cite{DBLP:conf/icml/JiangK022}}  & 53.81$\pm$0.58   & 4.08      & 44.23$\pm$0.74  & 6.07 & 45.92$\pm$0.76   & 2.91  \\ 
\multicolumn{2}{c}{GAP \cite{kang2023meta}}  & 55.12$\pm$0.61   & 4.83      & 46.54$\pm$0.71  & 7.24 & 48.26$\pm$0.69   & 3.65  \\\midrule
\multirow{4}{*}{\begin{tabular}[c]{@{}c@{}}GC\\Scratch \\ \cite{yong2020gradient}\end{tabular}} &1 epoch     & 23.01$\pm$1.02          & 30.79   & 21.44$\pm$1.02         & 60.99     & 18.67$\pm$0.34   & 4.01\\
&5 epochs     & 49.77$\pm$0.54 & 155.34  & 34.41$\pm$0.78    & 310.33   & 24.43$\pm$0.83   & 25.87  \\
&10 epochs    & 68.56$\pm$0.39 & 310.29  & 49.89$\pm$0.53    & 622.01   & 35.49$\pm$0.61   & 51.70 \\
&20 epochs    & 75.04$\pm$0.61 & 623.33 & 66.78$\pm$0.47    & 1248.98   & 44.05$\pm$0.52   & 102.63\\  
&50 epochs   & 77.98$\pm$0.45  & 1557.76    & 82.03$\pm$0.42   & 3109.14  & 53.58$\pm$0.61   & 257.24\\ \midrule 
\multicolumn{2}{c}{PudNet} & \textbf{58.51}$\pm$0.63 & 0.51     & \textbf{51.03}$\pm$0.68   & 0.63 & \textbf{54.50}$\pm$0.74   & 0.32 \\ \bottomrule
\end{tabular} 
\label{IN-1K-cross}
\end{table*}

\begin{table*}[h]
\setlength\tabcolsep{2pt}
\renewcommand\arraystretch{1}
\centering
\small
\vspace{-0.02in}
\caption{Ablation study of our method PudNet. `IN' denotes `ImageNet-100'.}
\label{ablation}
\begin{tabular}{@{}ccccccc@{}}
\toprule
\multirow{2}{*}{\begin{tabular}[c]{@{}c@{}}Method\end{tabular}}   & \multicolumn{3}{c}{Intra-dataset}   & \multicolumn{3}{c}{Inter-dataset}\\ 
            & Fashion-set     & CIFAR-100-set    & ImageNet-100-set  & IN$\to$Animals-10 & IN$\to$CIFAR-10   & IN$\to$DTD   \\ \midrule
PudNet-w.o.-Context & 93.08$\pm$0.44  & 65.35$\pm$0.51  & 61.42$\pm$0.70         & 37.21$\pm$0.51   & 39.43$\pm$0.68    & 37.79$\pm$0.83      \\ \midrule
PudNet-metric       & 94.75$\pm$0.44  & 68.60$\pm$0.61  & 61.28$\pm$0.85  & 40.35$\pm$0.60   & 43.78$\pm$0.69    & 43.67$\pm$0.72  \\
PudNet-w.o.-kl      & 95.44$\pm$0.38  & 70.27$\pm$0.54  & 67.53$\pm$0.73  & 41.23$\pm$0.68   & 43.98$\pm$0.74    & 45.12$\pm$0.88  \\ \midrule
PudNet              & \textbf{96.24}$\pm$0.39 & \textbf{73.33}$\pm$0.54 & \textbf{71.57}$\pm$0.71  & \textbf{42.43}$\pm$0.58  &\textbf{45.07}$\pm$0.70 & \textbf{47.50}$\pm$0.71  \\ \bottomrule
\end{tabular}
\label{abla}
\end{table*} 

\begin{table}[h]
\small
\setlength\tabcolsep{2.4pt}
\renewcommand\arraystretch{1}
\centering
\vspace{-0.1in}
\caption{Results of different methods in terms of different target network architectures on CIFAR-100-set.}
 \setlength{\tabcolsep}{0.6mm}
 {
\begin{tabular}{@{}cccccc@{}}
\toprule
\multicolumn{2}{c}{} & \multicolumn{2}{c}{ConvNet-3}   & \multicolumn{2}{c}{ResNet-34}\\ \midrule
\multicolumn{2}{c}{Method}        & ACC(\%) & time(sec.) & ACC(\%) & time(sec.)\\ \midrule
\multicolumn{2}{c}{Pretrained}    & 58.35$\pm$0.61 & - & 65.03$\pm$0.53   & -\\ \midrule
\multicolumn{2}{c}{MatchNet \cite{vinyals2016matching}}      & 47.75$\pm$0.73 & - & 60.37$\pm$0.70   & -\\
\multicolumn{2}{c}{ProtoNet \cite{snell2017prototypical}}        & 51.96$\pm$0.57 & - & 64.28$\pm$0.58   & -\\
\multicolumn{2}{c}{Meta-baseline \cite{chen2021meta}} & 57.69$\pm$0.38 & - & 67.40$\pm$0.69   & -\\
\multicolumn{2}{c}{Meta-DeepDBC \cite{xie2022joint}}  & 60.52$\pm$0.41 & - & 69.64$\pm$0.75   & -\\
\multicolumn{2}{c}{MUSML \cite{DBLP:conf/icml/JiangK022}}         & 56.49$\pm$0.56 & 1.21 & 66.39$\pm$0.59   & 3.11 \\ \midrule
\multirow{3}{*}{\begin{tabular}[c]{@{}c@{}}Adam\\Scratch\\\cite{DBLP:journals/corr/KingmaB14}\end{tabular}} 
&1 epoch & 49.11$\pm$1.03 & 0.99 & 47.39$\pm$1.36   & 5.47    \\
&30 epochs & 64.54$\pm$0.40 & 28.37 & 71.17$\pm$0.53   & 153.87  \\
&50 epochs & 70.68$\pm$0.53 & 49.02 & 78.72$\pm$0.71   & 263.25    \\ \midrule
\multirow{3}{*}{\begin{tabular}[c]{@{}c@{}}GC\\Scratch \\ \cite{yong2020gradient} \end{tabular}} 
&1 epoch  & 50.23$\pm$1.23 & 0.99  & 48.44$\pm$1.41          & 5.52  \\
&30 epochs  & 66.76$\pm$0.54 & 29.8 & 72.37$\pm$0.75  & 154.19 \\
&50 epochs  & 71.56$\pm$0.63 & 50.2 & 79.85$\pm$0.83  & 264.03 \\ \midrule 
\multicolumn{2}{c}{PudNet}    & \textbf{64.09}$\pm$0.40 & 0.03   & \textbf{72.87}$\pm$0.64   & 0.59\\ \bottomrule
\end{tabular}
}
\label{Res34}
\vspace{-0.1in}
\end{table}

\textbf{Performance Analysis of Inter-dataset Prediction.} 
We further evaluate the performance of our method in the inter-dataset setting. 
The results are listed in Table \ref{cifar-inter-res18} and Table \ref{mini-cross}, highlighting the efficiency of our PudNet. For instance, as shown in Table \ref{mini-cross}, our PudNet can achieve a comparable top-1 accuracy with that of training the model from scratch on CIFAR-10 at around 10 epochs on the ImageNet-100$\to$CIFAR-10 dataset, while PudNet is at least 1000 times faster than traditional training methods, Adam and GC. 
Besides, our model still outperforms the meta-learning methods by a large margin.
More surprisingly, although the label space of DTD (a texture classification dataset) is significantly different from that of ImageNet-100, our method still demonstrates good efficacy.
We expect that these results can motivate more researchers to explore this direction. 

\begin{table}
\setlength\tabcolsep{6pt}
\renewcommand\arraystretch{1}
\small
\centering
\vspace{-0.1in}
\caption{Performance of finetuning all methods on target network ResNet-18 using 50 epochs.}
\label{ft}
 \setlength{\tabcolsep}{0.7mm}
 {
\begin{tabular}{@{}cccc@{}}
\toprule
Method       & CIFAR-100-set   & ImageNet-100$\to$DTD\\ \midrule
Pretrained   & 59.75$\pm$0.24 & 53.43$\pm$0.43\\ \midrule
MatchNet \cite{vinyals2016matching}     & 56.83$\pm$0.40 & 52.26$\pm$0.71\\
ProtoNet \cite{snell2017prototypical}      & 58.14$\pm$0.29 & 54.44$\pm$0.64\\
Meta-baseline \cite{chen2021meta}& 60.05$\pm$0.27 & 54.52$\pm$0.58\\
Meta-DeepDBC \cite{xie2022joint}& 61.35$\pm$0.28 & 55.09$\pm$0.63\\ 
MUSML \cite{DBLP:conf/icml/JiangK022}       & 59.28$\pm$0.29 & 53.61$\pm$0.69\\ \midrule
Adam Scratch \cite{DBLP:journals/corr/KingmaB14} & 48.25$\pm$0.30 & 52.16$\pm$0.56\\ 
GC Scratch \cite{yong2020gradient}   & 50.03$\pm$0.32 & 53.58$\pm$0.61\\ \midrule
PudNet       & \textbf{65.19}$\pm$0.22 & \textbf{58.25}$\pm$0.57\\ \bottomrule
\end{tabular}
}
\label{finetune}
\vspace{-0.1in}
\end{table}


\textbf{Scaling Up to ImageNet-1K.}
For the intra-dataset setting, we partition ImageNet-1K into two mutually exclusive sets of 800 classes for training and 200 classes for testing.
The top-5 accuracies of different methods are reported in Table \ref{IN-1K-intra}. We can observe that our method still achieves surprisingly good efficiency. For instance, It takes 8,967 GPU seconds to train ResNet-18 on ImageNet-1K using GC from scratch and obtains a top-5 accuracy of 44.65\%. However, our PudNet only costs 3.89 GPU seconds to predict the network parameters of ResNet-18 achieving a comparable performance (44.92\%), more than 2,300 times faster than the traditional training paradigm. In addition, our method also outperforms the state-of-the-art meta-learning methods by a large margin.
These results highlight the effective scalability of PudNet to larger and more complex datasets.

For the inter-dataset setting, we conduct additional experiments on three cross-domain datasets: ImageNet-1K $\to$ Animals-10, ImageNet-1K $\to$ CIFAR-10, and ImageNet-1K $\to$ DTD. The results are presented in Table \ref{IN-1K-cross}. 
Remarkably, our method continues to demonstrate impressive efficiency.
Please kindly note that compared to our PudNet trained on ImageNet-100, as in Table \ref{mini-cross}, which achieves 42.43\% top-1 accuracy on Animals-10, 45.07\% top-1 accuracy on CIFAR-10, and 47.5\% top-1 accuracy on DTD, training PudNet on the larger ImageNet dataset yields an improvement of approximately 16\%, 6\%, and 7\% on the three inter-dataset tasks, respectively, while requiring comparable time costs. These results indicate that training our model with more complex and comprehensive datasets results in better generalization performance.
Based on the experiments conducted in both the intra-dataset and inter-dataset settings, it is evident that our PudNet can effectively scale to larger datasets.

\begin{figure*}[t]
\centering
\subfloat[Fashion-set]{\includegraphics[width=0.26\linewidth]{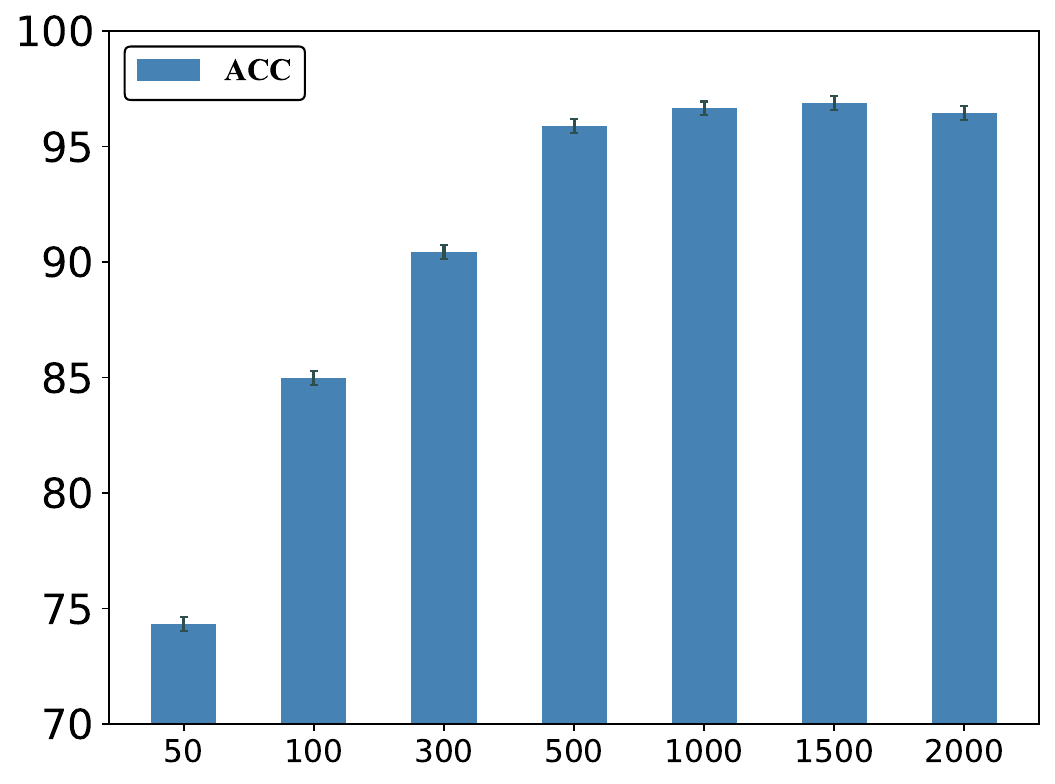}}  
\hfil
\subfloat[CIFAR-100-set]{\includegraphics[width=0.25\linewidth]{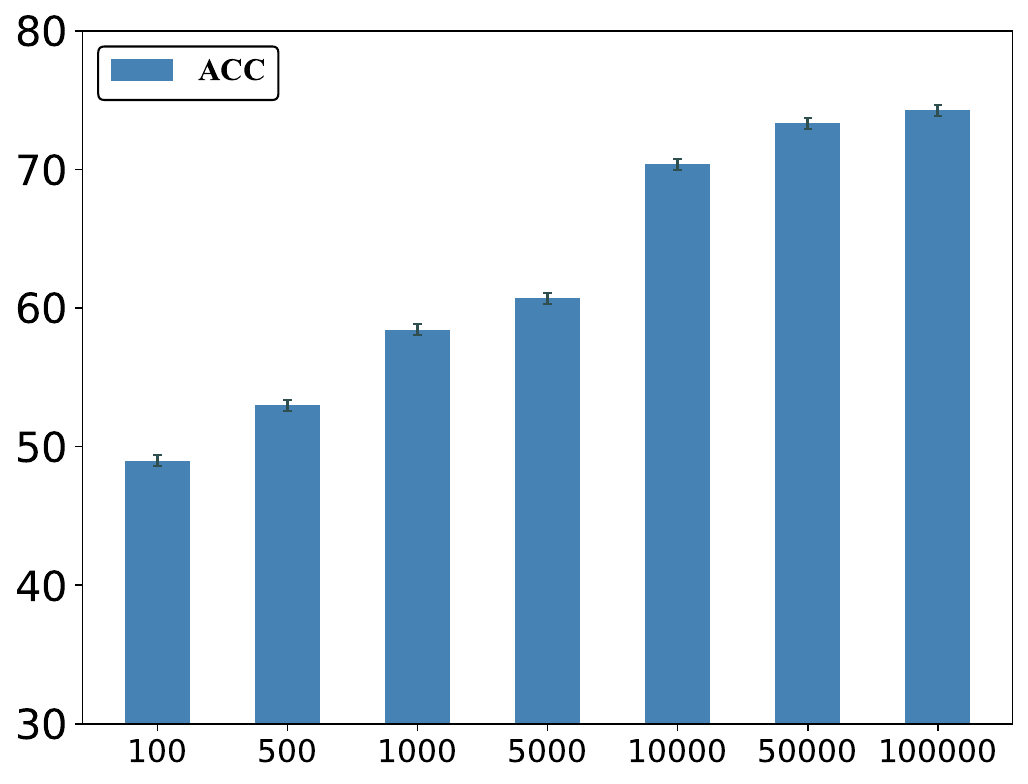}}
\hfil
\subfloat[ImageNet-100-set]{\includegraphics[width=0.25\linewidth]{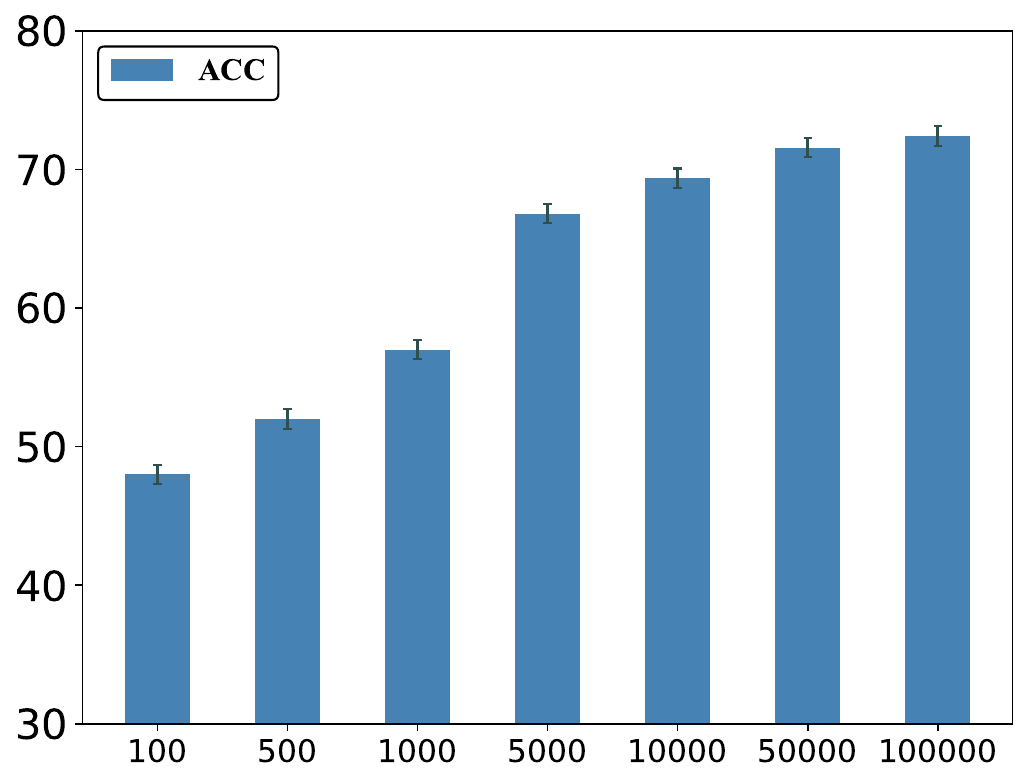}} 
\hfil
\caption{Effect of different groups of datasets in terms of the target network ResNet-18.}
\label{group_res18}
\vspace{-0.105in} 
\end{figure*}

\begin{figure*}[!h]
\centering
\subfloat[Fashion-set with ResNet-18]{\includegraphics[width=0.28\linewidth]{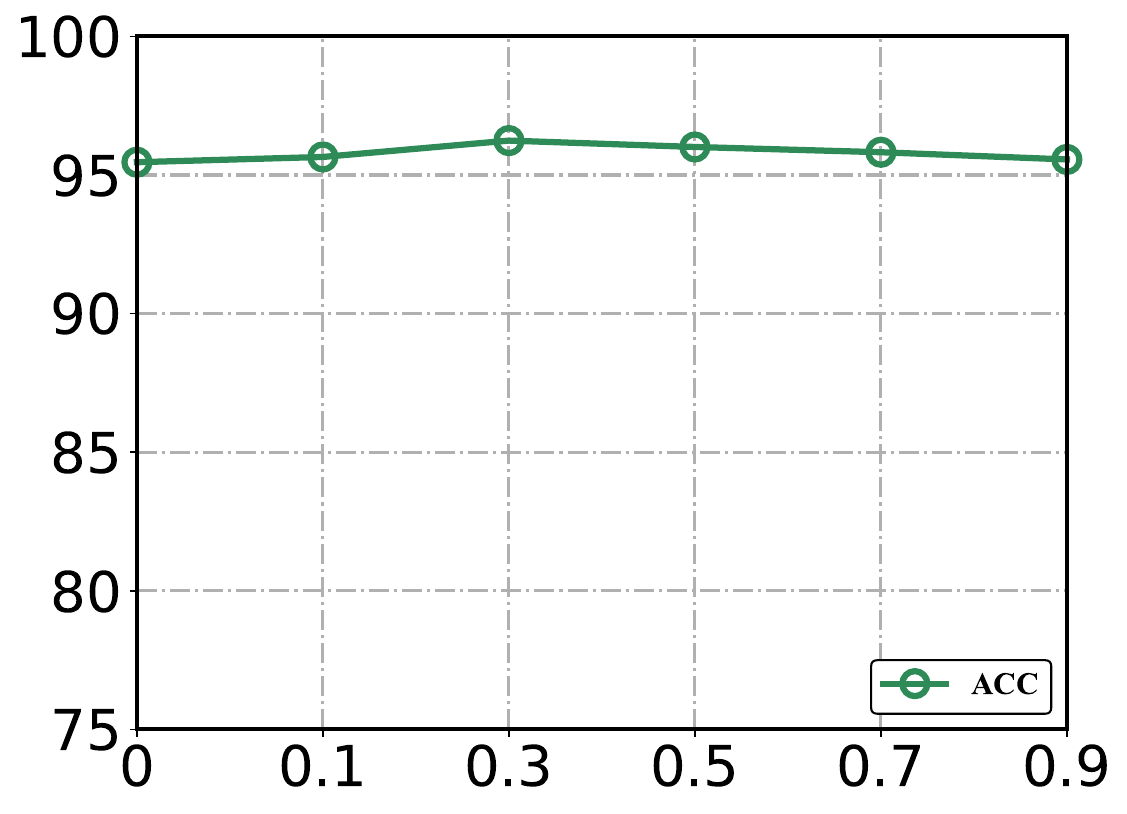}} \hfil 
\subfloat[\!CIFAR-100-set\! with ResNet-18\!]{\includegraphics[width=0.28\linewidth]{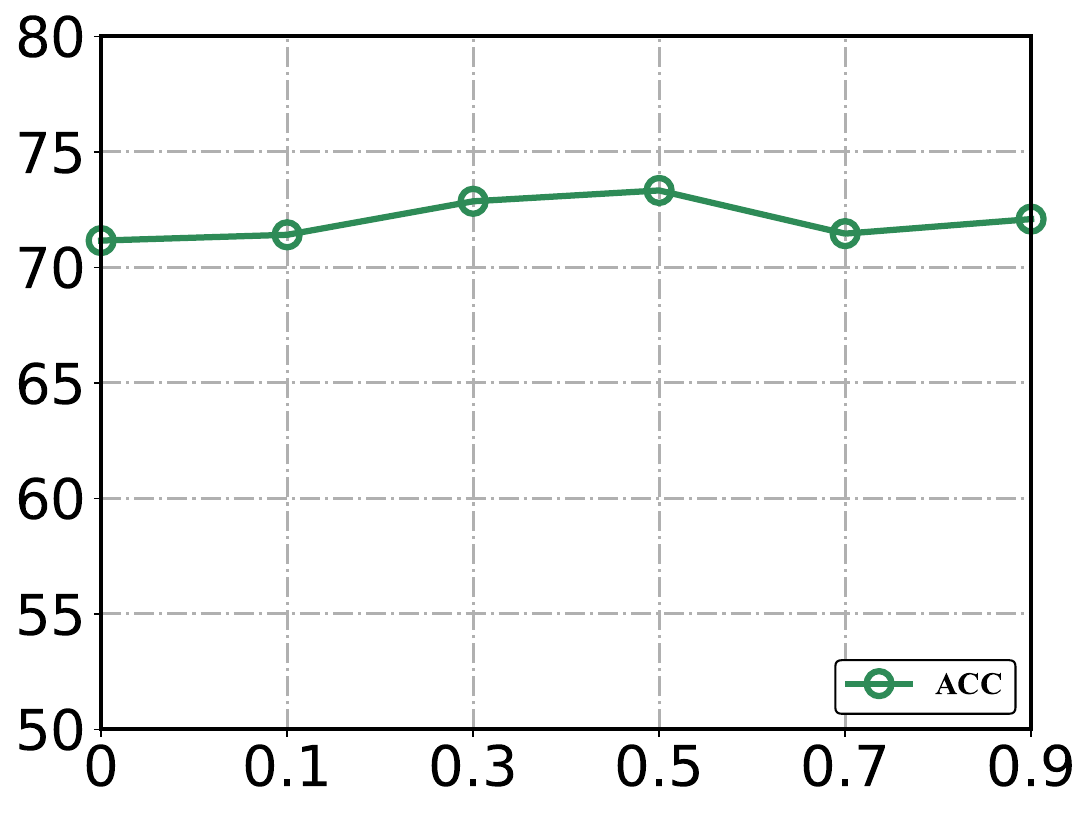}} \hfil
\subfloat[ImageNet-100-set with ResNet-18]{\includegraphics[width=0.28\linewidth]{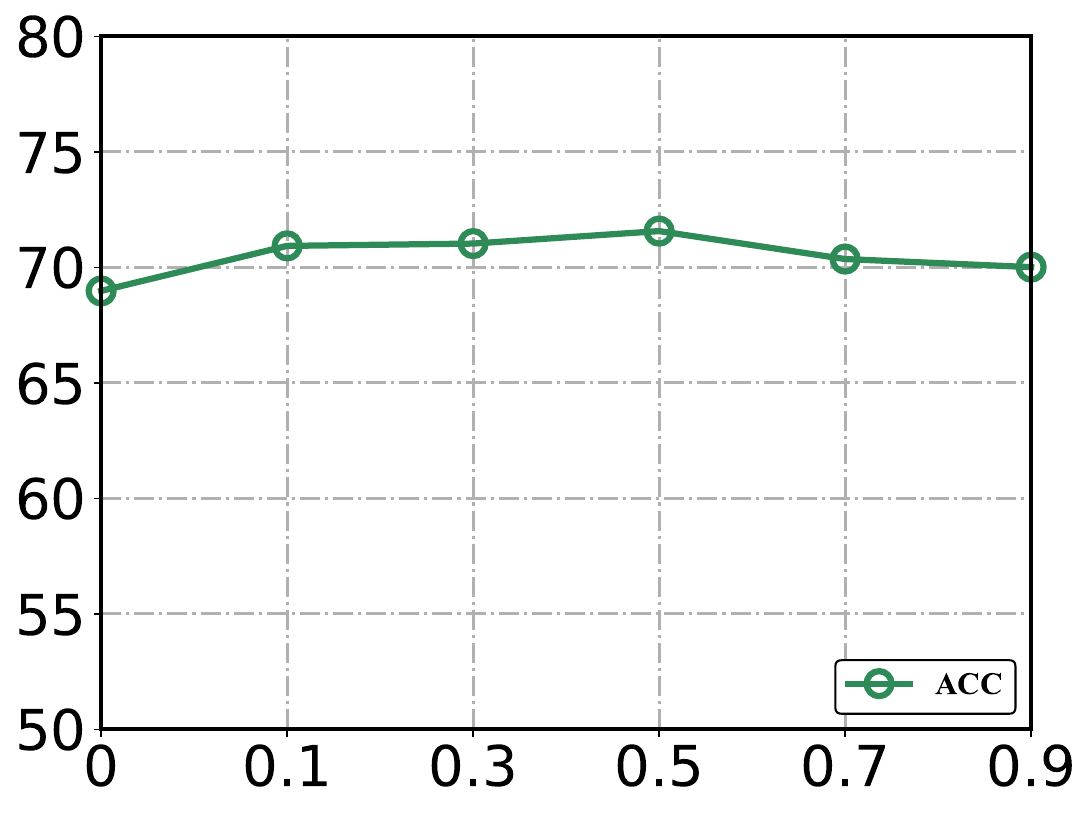}}
\caption{Sensitivity analysis of hyper-parameter $\eta$.}
\label{sensity}
\vspace{-0.2in}
\end{figure*}

\begin{figure*}[!h]
\centering
\subfloat[metric loss on Fashion-set]{\includegraphics[width=0.30\linewidth]{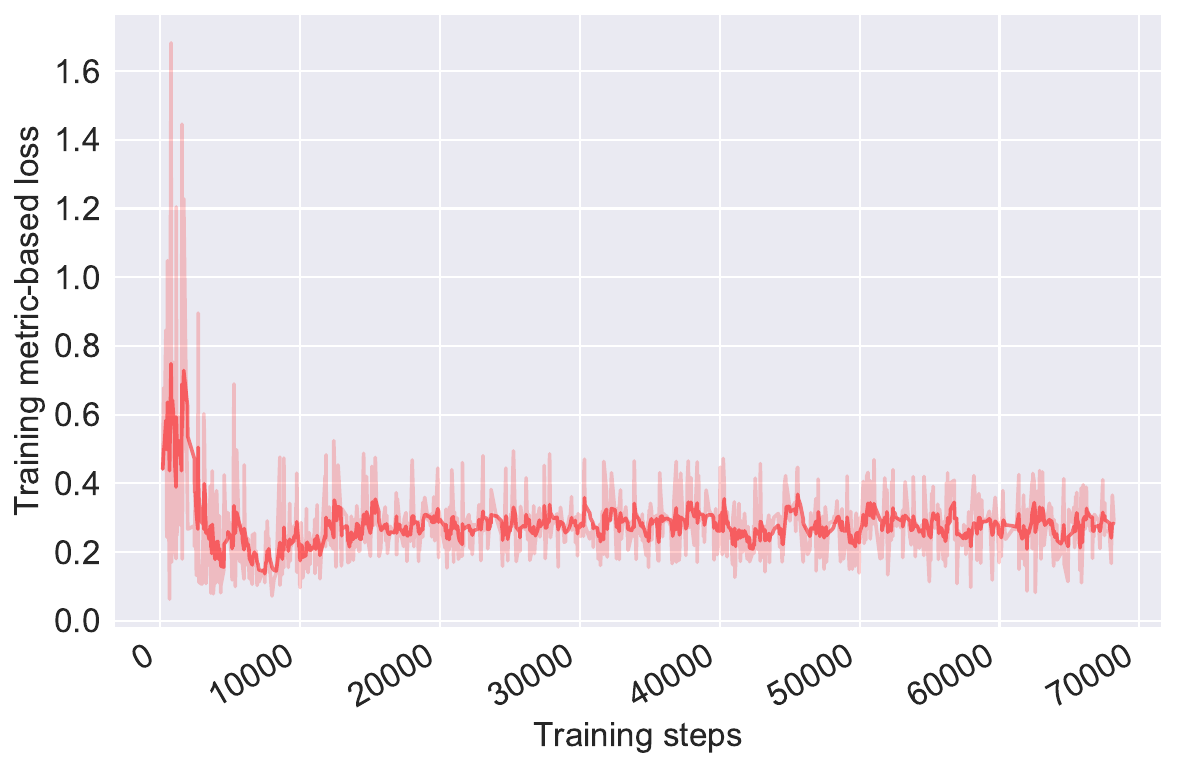}}
\hfil 
\subfloat[metric loss on CIFAR-100-set]{\includegraphics[width=0.30\linewidth]{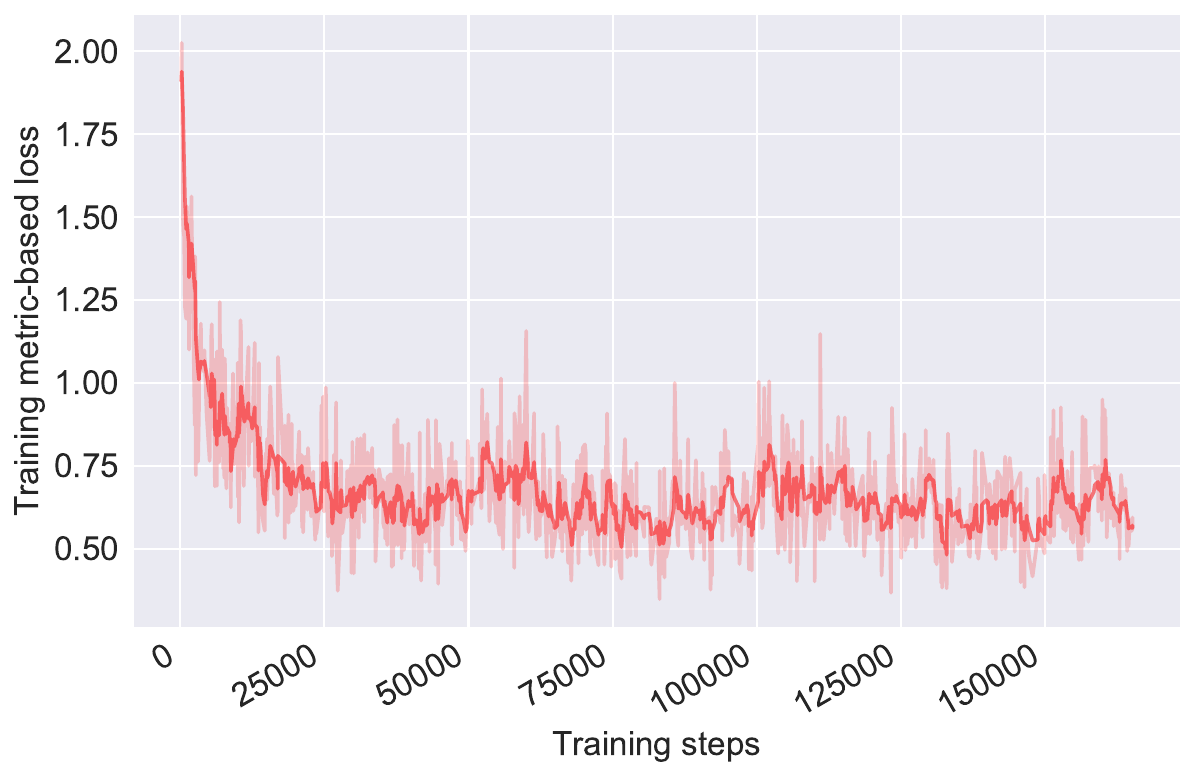}}
\hfil
\subfloat[metric loss on ImageNet-100-set]{\includegraphics[width=0.30\linewidth]{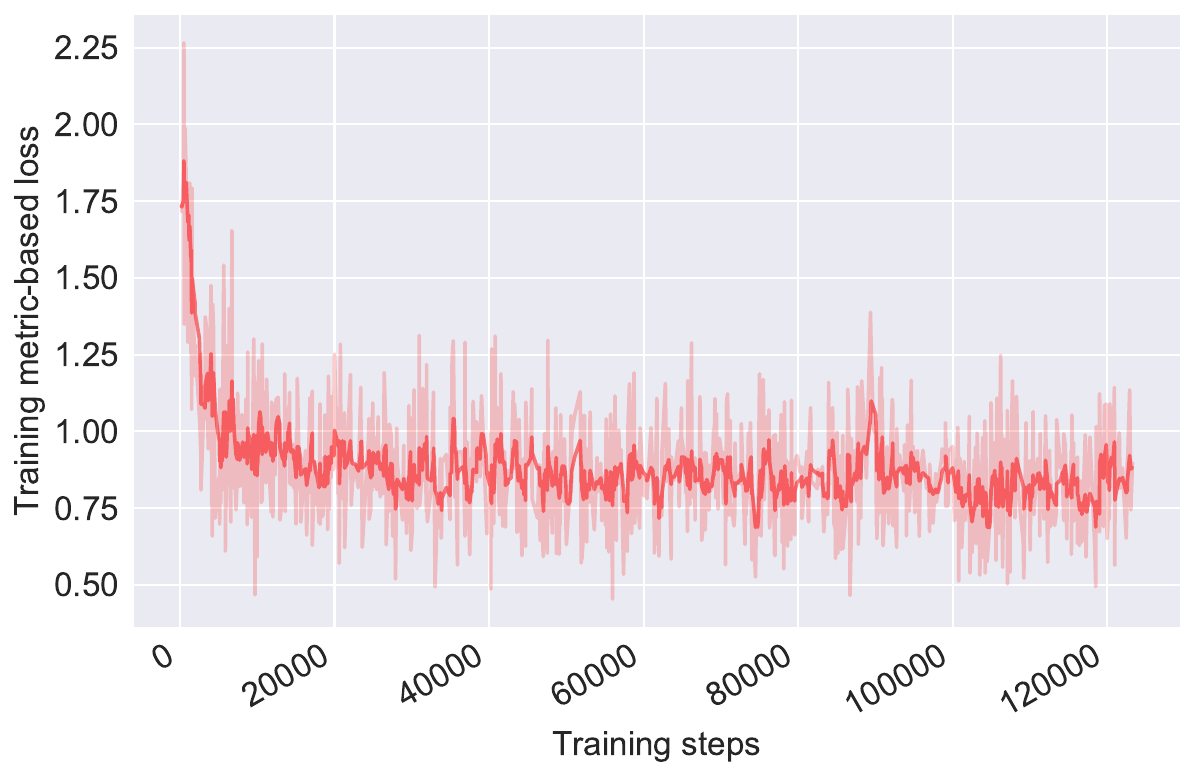}}
\hfil
\subfloat[total training loss on Fashion-set]{\includegraphics[width=0.30\linewidth]{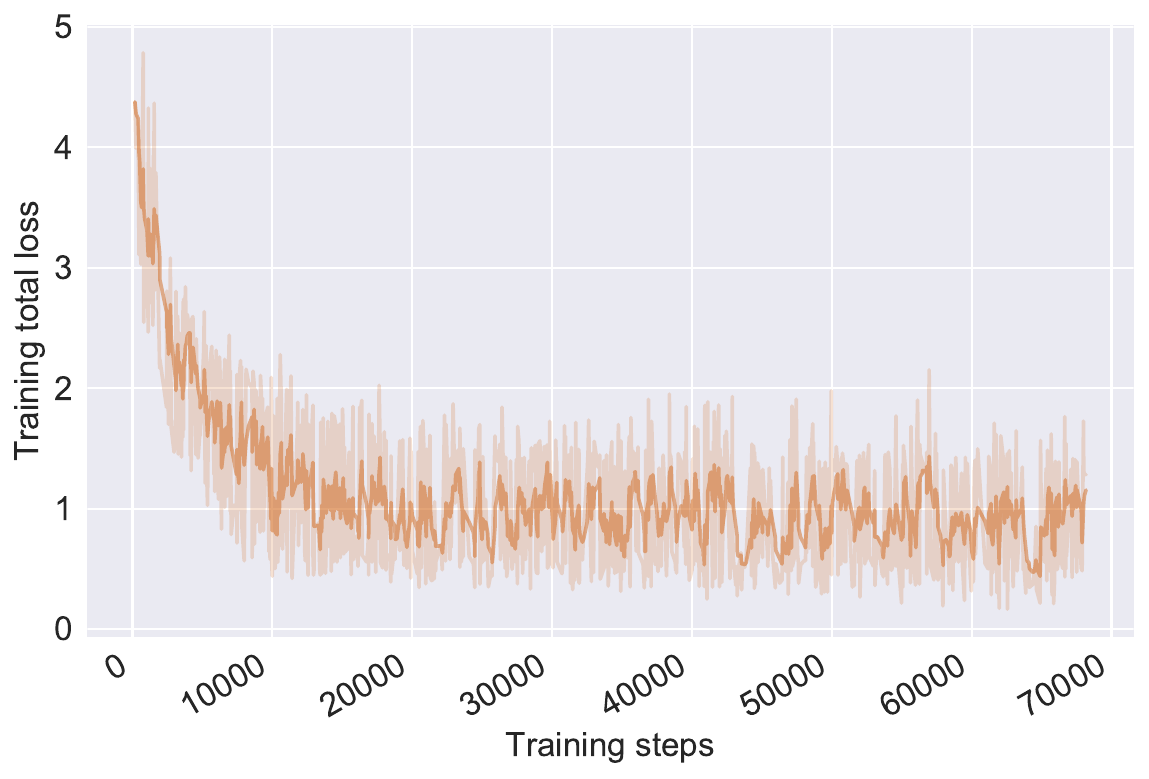}}
\hfil
\subfloat[total training loss on CIFAR-100-set]{\includegraphics[width=0.30\linewidth]{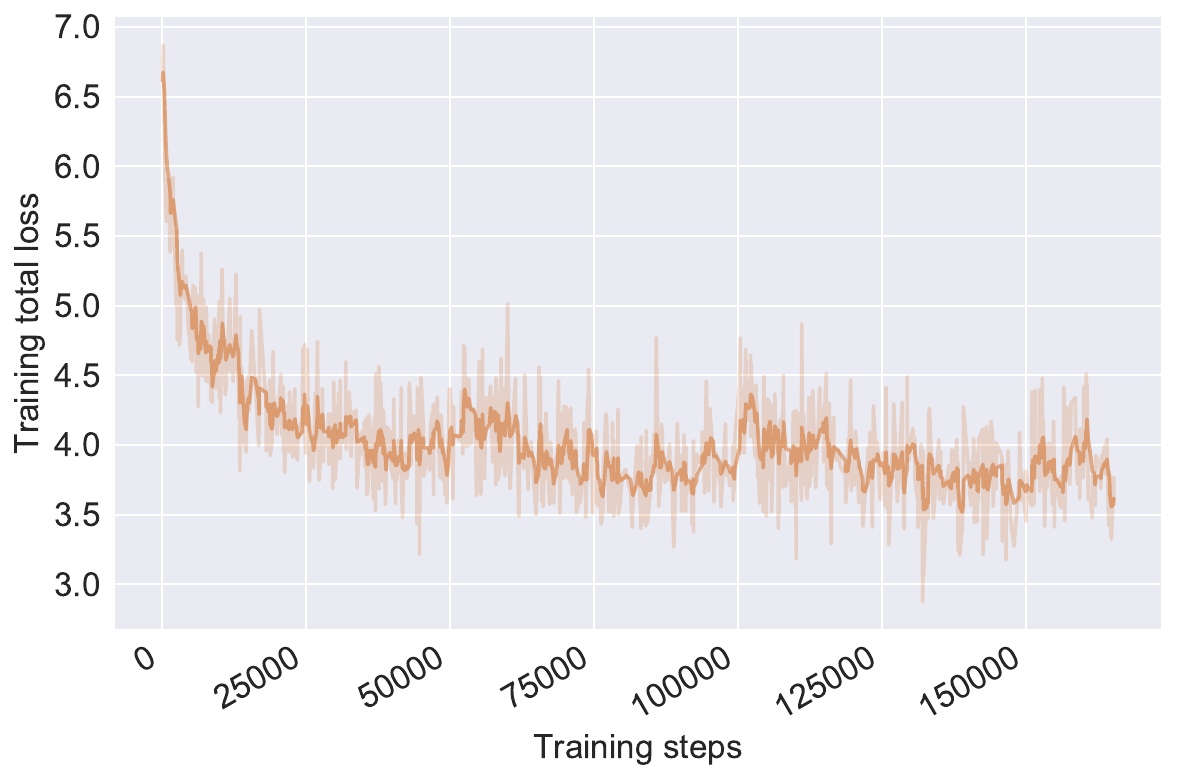}}
\hfil
\subfloat[total training loss on ImageNet-100-set]{\includegraphics[width=0.30\linewidth]{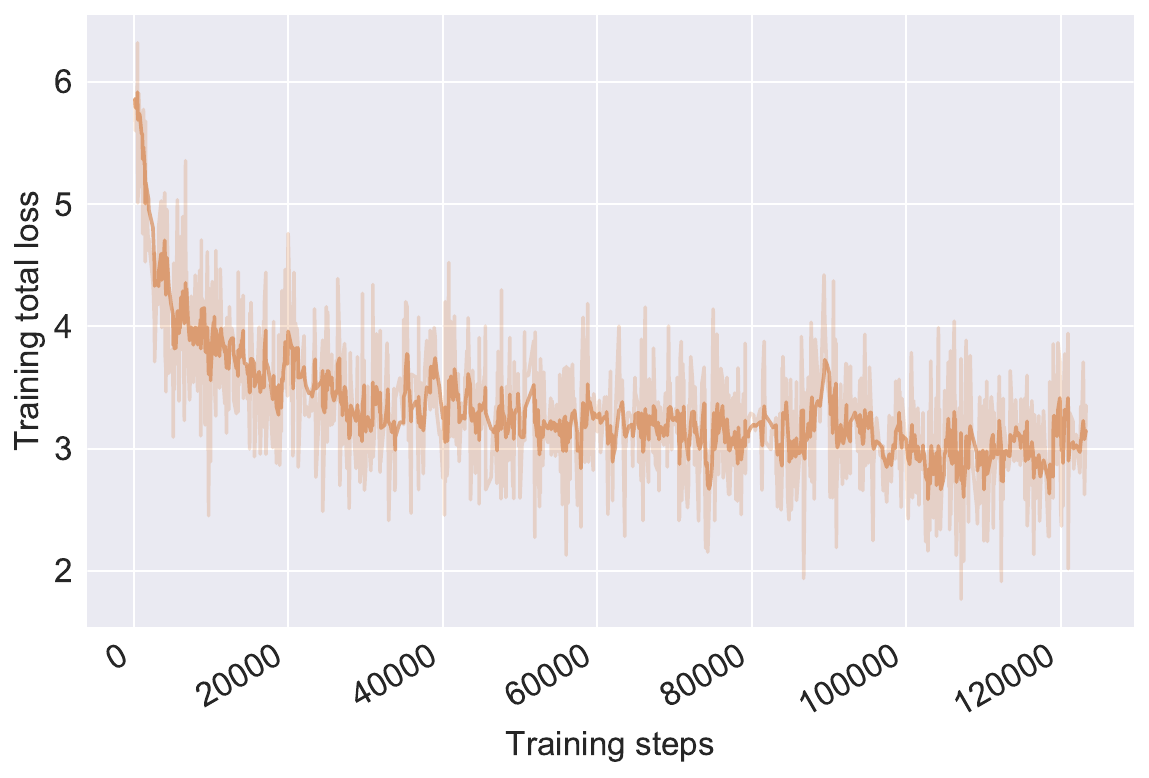}}
\caption{Training loss of PudNet.}
\label{train_loss}
\vspace{-0.1in}
\end{figure*}

\textbf{Ablation Study.}
We design several variants of our method to analyze the effect of different components in both intra-dataset and inter-dataset settings. The results are listed in Table \ref{abla}.
We first examine the effect of the context relation information. 
`PudNet-w.o.-Context' denotes that the dataset sketch is directly fed into the weight generator without utilizing AHRU.
Our PudNet outperforms `PudNet-w.o.-Context' by a significant margin, demonstrating the effectiveness of capturing dependencies among parameters of different layers.

We further evaluate the effect of the auxiliary task. `PudNet-metric' means our method only utilizes the parameter-free loss. `PudNet-w.o.-kl' means our method does not use the KL Divergence. As shown in Table \ref{abla}, `PudNet-w.o.-kl' achieves better performance compared to `PudNet-metric', highlighting the  effectiveness of the auxiliary full classification task. Moreover, PudNet outperforms `PudNet-w.o.-kl', thus illustrating the effectiveness of matching the predicted probability distributions of the parameter-free classification and full classification method.

\textbf{Performance Analysis of Different Target Networks.}
We conduct experiments to further evaluate our method by predicting parameters for different target network architectures, namely ConvNet-3 and ResNet-34.
The results are presented in Table \ref{Res34}.
We observe that our method can achieve comparable top-1 accuracy to those of GC at 30 epochs for both target network architectures, while our method exhibits more than 250 times faster than GC. This further demonstrates the efficiency of our method.



\begin{table*}[h]
\centering
\caption{Results of different methods on ImageNet-1K-set in the Intra-dataset setting for image denoising}
\begin{tabular}{@{}ccccccccc@{}}
\toprule
\multicolumn{2}{c}{\multirow{2}{*}{Method}}                                          & \multirow{2}{*}{time(sec.)}                   & \multicolumn{2}{c}{Gaussian ($\sigma=0.25$)}                      & \multicolumn{2}{c}{Gaussian ($\sigma=0.50$)}                      & \multicolumn{2}{c}{Gaussian ($\sigma=0.75$)}                      \\
\multicolumn{2}{c}{}                                                           &          & PNSR(dB)                        & SSIM                            & PNSR(dB)                        & SSIM                            & PNSR(dB)                        & SSIM                            \\ \midrule
\multirow{4}{*}{\begin{tabular}[c]{@{}c@{}}Adam\\ Scratch\end{tabular}} & 1 epoch    & $0.13 \times 10^3$ & 21.69                           & 0.871                           & 20.88                           & 0.851                           & 18.98                           & 0.814                           \\
                                                                        & 50 epochs  & $6.72 \times 10^3$ & 28.01                           & 0.947                           & 25.84                           & 0.919                           & 24.21                           & 0.891                           \\
                                                                        & 100 epochs & $1.36 \times 10^4$ & 28.45                           & 0.951                           & 26.25                           & 0.920                           & 24.65                           & 0.893                           \\
                                                                        & 300 epochs & $4.17 \times 10^4$ & 29.56                           & 0.956                           & 26.78                           & 0.927                           & 25.02                           & 0.902                           \\ \midrule
\multirow{4}{*}{\begin{tabular}[c]{@{}c@{}}GC\\ Scratch\end{tabular}}   & 1 epoch    & $0.16\times 10^3$  & 21.85                           & 0.869                           & 21.01                           & 0.846                           & 18.71                           & 0.821                           \\
                                                                        & 50 epochs  & $6.80 \times 10^3$ & 28.47                           & 0.947                           & 25.95                           & 0.919                           & 24.64                           & 0.892                           \\
                                                                        & 100 epochs & $1.37 \times 10^4$ & 28.64                           & 0.953                           & 26.92                           & 0.924                           & 25.02                           & 0.901                           \\
                                                                        & 300 epochs & $4.17 \times 10^4$ & 30.12                           & 0.959                           & 27.23                           & 0.933                           & 25.77                           & 0.914                           \\ \midrule
\multicolumn{2}{c}{PudNet}                                                           & 0.91              & \textbf{28.53} & \textbf{0.950} & \textbf{25.97} & \textbf{0.921} & \textbf{24.52} & \textbf{0.892} \\ \bottomrule
\end{tabular}
\label{denoisy_intra}
\vspace{-0.06in}
\end{table*}


\begin{table*}[h]
\centering
\caption{Results of different methods in the Inter-dataset setting for image denoising}
\begin{tabular}{@{}cccccccccccc@{}}
\toprule
\multirow{2}{*}{Noise Type}                                                         & \multicolumn{2}{c}{\multirow{2}{*}{Method}}                                        & \multicolumn{3}{c}{ImageNet-1K$\to$Animals-10}                                         & \multicolumn{3}{c}{ImageNet-1K$\to$CIFAR-10}                                           & \multicolumn{3}{c}{ImageNet-1K$\to$DTD}                                                \\
                                                                                    & \multicolumn{2}{c}{}                                                               & PNSR(dB)                        & SSIM                            & time(sec.)         & PNSR(dB)                        & SSIM                            & time(sec.)         & PNSR(dB)                        & SSIM                            & time(sec.)         \\ \midrule
\multirow{5}{*}{\begin{tabular}[c]{@{}c@{}}Gaussian\\ ($\sigma=0.25$)\end{tabular}} & \multirow{4}{*}{\begin{tabular}[c]{@{}c@{}}GC\\ Scratch\end{tabular}} & 1 epoch    & 21.52                           & 0.891                           & $0.14 \times 10^3$ & 22.56                           & 0.902                           & $0.11 \times 10^3$ & 21.04                           & 0.723                           & $0.57 \times 10^2$ \\
                                                                                    &                                                                       & 50 epochs  & 27.28                           & 0.957                           & $7.30 \times 10^3$ & 32.58                           & 0.973                           & $5.34 \times 10^3$ & 26.18                           & 0.809                           & $2.86 \times 10^3$ \\
                                                                                    &                                                                       & 100 epochs & 28.87                           & 0.963                           & $1.45 \times 10^4$ & 34.12                           & 0.980                           & $1.08 \times 10^4$ & 27.43                           & 0.834                           & $5.80 \times 10^3$ \\
                                                                                    &                                                                       & 300 epochs & 29.51                           & 0.968                           & $4.09 \times 10^4$ & 36.01                           & 0.984                           & $3.20 \times 10^4$ & 29.01                           & 0.851                           & $1.74 \times 10^4$ \\ \cmidrule(l){2-12} 
                                                                                    & \multicolumn{2}{c}{PudNet}                                                         & \textbf{28.32} & \textbf{0.960} & 1.48               & \textbf{33.52} & \textbf{0.975} & 1.14               & \textbf{27.66} & \textbf{0.819} & 1.45               \\ \midrule
\multirow{5}{*}{\begin{tabular}[c]{@{}c@{}}Gaussian\\ ($\sigma=0.50$)\end{tabular}} & \multirow{4}{*}{\begin{tabular}[c]{@{}c@{}}GC\\ Scratch\end{tabular}} & 1 epoch    & 20.40                           & 0.872                           & $0.14 \times 10^3$ & 21.65                           & 0.887                           & $0.10 \times 10^3$ & 18.85                           & 0.587                           & $0.57 \times 10^2$ \\
                                                                                    &                                                                       & 50 epochs  & 24.51                           & 0.919                           & $7.30 \times 10^3$ & 30.53                           & 0.961                           & $5.33 \times 10^3$ & 23.40                           & 0.703                           & $2.87 \times 10^3$ \\
                                                                                    &                                                                       & 100 epochs & 25.64                           & 0.931                           & $1.45 \times 10^4$ & 31.28                           & 0.964                           & $1.08 \times 10^4$ & 24.79                           & 0.740                           & $5.81 \times 10^3$ \\
                                                                                    &                                                                       & 300 epochs & 26.55                           & 0.941                           & $4.09 \times 10^4$ & 32.14                           & 0.971                           & $3.20 \times 10^4$ & 26.22                           & 0.762                           & $1.74 \times 10^4$ \\ \cmidrule(l){2-12} 
                                                                                    & \multicolumn{2}{c}{PudNet}                                                         & \textbf{25.49} & \textbf{0.930} & 1.49               & \textbf{30.82} & \textbf{0.960} & 1.16               & \textbf{25.39} & \textbf{0.716} & 1.44               \\ \midrule
\multirow{5}{*}{\begin{tabular}[c]{@{}c@{}}Gaussian\\ ($\sigma=0.75$)\end{tabular}} & \multirow{4}{*}{\begin{tabular}[c]{@{}c@{}}GC\\ Scratch\end{tabular}} & 1 epoch    & 19.58                           & 0.852                           & $0.15 \times 10^3$ & 21.35                           & 0.875                           & $0.10 \times 10^3$ & 15.97                           & 0.475                           & $0.57 \times 10^2$ \\
                                                                                    &                                                                       & 50 epochs  & 23.35                           & 0.893                           & $7.30 \times 10^3$ & 28.48                           & 0.944                           & $5.34 \times 10^3$ & 21.67                           & 0.632                           & $2.88 \times 10^3$ \\
                                                                                    &                                                                       & 100 epochs & 24.21                           & 0.908                           & $1.45 \times 10^4$ & 29.47                           & 0.951                           & $1.08 \times 10^4$ & 23.33                           & 0.651                           & $5.82 \times 10^3$ \\
                                                                                    &                                                                       & 300 epochs & 24.65                           & 0.914                           & $4.09 \times 10^4$ & 29.53                           & 0.954                           & $3.20 \times 10^4$ & 24.47                           & 0.675                           & $1.74 \times 10^4$ \\ \cmidrule(l){2-12} 
                                                                                    & \multicolumn{2}{c}{PudNet}                                                         & \textbf{24.30} & \textbf{0.893} & 1.49               & \textbf{28.43} & \textbf{0.947} & 1.17               & \textbf{23.83} & \textbf{0.636} & 1.45               \\ \bottomrule
\end{tabular}
\label{denoisy_inter1}
\vspace{-0.1in}
\end{table*}

\textbf{Finetuning Predicted Parameters.} 
Since a typical strategy for applying a pretrained model to a new dataset is to finetune the model, we evaluate our method by finetuning the models obtained from both the intra-dataset and the inter-dataset settings.  
To facilitate this evaluation, we incorporate an additional linear classification layer into PudNet, as well as and baselines (excluding `Adam Scratch' and `GC Scratch'). 
Subsequently, we randomly select 10,000 images from CIFAR-100 and 800 images from DTD for finetuning the respective models, while the remaining images are reserved for testing. The results displayed in Table \ref{finetune} demonstrate that our method outperforms other competitors and achieves the best performance.
This indicates that the predicted parameters generated by PudNet can effectively serve as a pretrained model. 


\textbf{Effect of Different Groups of Datasets.} 
We analyze the effect of different groups of datasets on training PudNet.
Fig. \ref{group_res18} reports the results obtained by exploiting PudNet to predict parameters for ResNet-18. 
We construct varying groups of datasets using Fashion-set, CIFAR-100-set, and ImageNet-100-set, respectively. 
We find that with more groups of datasets for training, our PudNet achieves better performance. 
This improvement can be attributed to the fact that a larger number of groups provide more opportunities for our PudNet to learn the hyper-mapping relation, thereby enhancing its generalization ability. 
However, as the number of groups becomes large, the performance increase becomes slow.


\textbf{Parameter Sensitive Analysis.}
We analyze the effect of different values of the hyper-parameter $\eta$.
Recall that $\eta$ controls the percent of dataset complementary information in the initial residual connection. 
Fig. \ref{sensity}(a)(b)(c) show the results in terms of ResNet-18. We observe that our model obtain better performance when $\eta > 0$ in general.
Additionally, our method is not sensitive to $\eta$ in a relatively large range.


\textbf{Convergence Analysis.} 
We discuss the convergence property of the proposed method by plotting the loss curves with increasing iteration. Here we utilize PudNet to predict parameters for ResNet-18, based on Fashion-set, CIFAR-100-set and ImageNet-100-set respectively. 
As shown in Fig.~\ref{train_loss}, the training metric-based loss (abbreviated as metric loss) and training total loss first decrease rapidly as the number of iterations increases, and then gradually decrease to convergence.


\begin{table*}[!h]
\centering
\caption{Results of different methods in the Inter-dataset setting for image denoising}
\begin{tabular}{@{}cccccccccccc@{}}
\toprule
\multirow{2}{*}{Noise Type}                                                         & \multicolumn{2}{c}{\multirow{2}{*}{Method}}                                        & \multicolumn{3}{c}{ImageNet-1K$\to$CIFAR-100}                                          & \multicolumn{3}{c}{ImageNet-1K$\to$CUB-200}                                            & \multicolumn{3}{c}{ImageNet-1K$\to$BSDS500}                                            \\
                                                                                    & \multicolumn{2}{c}{}                                                               & PNSR(dB)                        & SSIM                            & time(sec.)         & PNSR(dB)                        & SSIM                            & time(sec.)         & PNSR(dB)                        & SSIM                            & time(sec.)         \\ \midrule
\multirow{5}{*}{\begin{tabular}[c]{@{}c@{}}Gaussian\\ ($\sigma=0.25$)\end{tabular}} & \multirow{4}{*}{\begin{tabular}[c]{@{}c@{}}GC\\ Scratch\end{tabular}} & 1 epoch    & 26.88                           & 0.933                           & $0.12 \times 10^3$ & 19.89                           & 0.826                           & $0.71 \times 10^2$ & 13.69                           & 0.443                           & $0.16 \times 10^2$ \\
                                                                                    &                                                                       & 50 epochs  & 32.24                           & 0.963                           & $5.82 \times 10^3$ & 25.92                           & 0.913                           & $3.57 \times 10^3$ & 28.29                           & 0.921                           & $0.84 \times 10^3$ \\
                                                                                    &                                                                       & 100 epochs & 34.13                           & 0.973                           & $1.19 \times 10^4$ & 26.75                           & 0.924                           & $7.25 \times 10^3$ & 30.55                           & 0.933                           & $1.69 \times 10^3$ \\
                                                                                    &                                                                       & 300 epochs & 35.65                           & 0.982                           & $2.99 \times 10^4$ & 27.61                           & 0.935                           & $2.21 \times 10^4$ & 32.84                           & 0.947                           & $5.07 \times 10^3$ \\ \cmidrule(l){2-12} 
                                                                                    & \multicolumn{2}{c}{PudNet}                                                         & \textbf{33.61} & \textbf{0.971} & 1.02               & \textbf{26.24} & \textbf{0.920} & 1.53               & \textbf{30.80} & \textbf{0.931} & 0.67               \\ \midrule
\multirow{5}{*}{\begin{tabular}[c]{@{}c@{}}Gaussian\\ ($\sigma=0.50$)\end{tabular}} & \multirow{4}{*}{\begin{tabular}[c]{@{}c@{}}GC\\ Scratch\end{tabular}} & 1 epoch    & 26.09                           & 0.912                           & $0.12 \times 10^3$ & 19.14                           & 0.794                           & $0.72 \times 10^2$ & 14.08                           & 0.439                           & $0.17 \times 10^2$ \\
                                                                                    &                                                                       & 50 epochs  & 30.42                           & 0.951                           & $5.83 \times 10^3$ & 23.81                           & 0.875                           & $3.58 \times 10^3$ & 25.44                           & 0.831                           & $0.85 \times 10^3$ \\
                                                                                    &                                                                       & 100 epochs & 31.26                           & 0.956                           & $1.19 \times 10^4$ & 24.57                           & 0.888                           & $7.25 \times 10^3$ & 26.50                           & 0.841                           & $1.70 \times 10^3$ \\
                                                                                    &                                                                       & 300 epochs & 32.27                           & 0.964                           & $3.00 \times 10^4$ & 25.14                           & 0.902                           & $2.21 \times 10^4$ & 28.59                           & 0.865                           & $5.07 \times 10^3$ \\ \cmidrule(l){2-12} 
                                                                                    & \multicolumn{2}{c}{PudNet}                                                         & \textbf{30.48} & \textbf{0.950} & 1.03               & \textbf{24.10} & \textbf{0.881} & 1.53               & \textbf{28.20} & \textbf{0.868} & 0.68               \\ \midrule
\multirow{5}{*}{\begin{tabular}[c]{@{}c@{}}Gaussian\\ ($\sigma=0.75$)\end{tabular}} & \multirow{4}{*}{\begin{tabular}[c]{@{}c@{}}GC\\ Scratch\end{tabular}} & 1 epoch    & 24.41                           & 0.887                           & $0.12 \times 10^3$ & 18.29                           & 0.763                           & $0.74 \times 10^2$ & 14.12                           & 0.369                           & $0.17 \times 10^2$ \\
                                                                                    &                                                                       & 50 epochs  & 28.75                           & 0.934                           & $5.83 \times 10^3$ & 22.41                           & 0.844                           & $3.59 \times 10^3$ & 22.73                           & 0.750                           & $0.85 \times 10^3$ \\
                                                                                    &                                                                       & 100 epochs & 29.17                           & 0.940                           & $1.20 \times 10^4$ & 23.09                           & 0.854                           & $7.26 \times 10^3$ & 23.75                           & 0.771                           & $1.70 \times 10^3$ \\
                                                                                    &                                                                       & 300 epochs & 29.92                           & 0.944                           & $3.00 \times 10^4$ & 23.60                           & 0.868                           & $2.21 \times 10^4$ & 25.83                           & 0.801                           & $5.07 \times 10^3$ \\ \cmidrule(l){2-12} 
                                                                                    & \multicolumn{2}{c}{PudNet}                                                         & \textbf{28.72} & \textbf{0.931} & 1.03               & \textbf{22.66} & \textbf{0.846} & 1.53               & \textbf{26.02} & \textbf{0.807} & 0.68               \\ \bottomrule
\end{tabular}
\label{denoisy_inter2}
\vspace{-0.06in}
\end{table*}

\begin{figure*}[h]
\centering
    \subfloat{\includegraphics[width=0.15\linewidth]{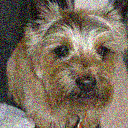}}
    \hfil
    \subfloat{\includegraphics[width=0.15\linewidth]{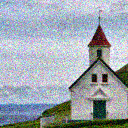}}
    \hfil
    \subfloat{\includegraphics[width=0.15\linewidth]{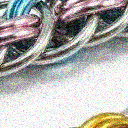}}
    \hfil
    \subfloat{\includegraphics[width=0.15\linewidth]{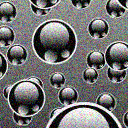}}
    \hfil
    \subfloat{\includegraphics[width=0.15\linewidth]{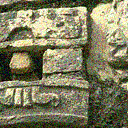}}
    \hfil
    \subfloat{\includegraphics[width=0.15\linewidth]{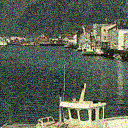}} 
    \hfil
    \subfloat {\includegraphics[width=0.15\linewidth]{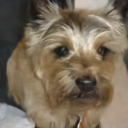}}
    \hfil
    \subfloat{\includegraphics[width=0.15\linewidth]{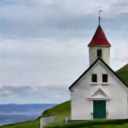}}
    \hfil
    \subfloat{\includegraphics[width=0.15\linewidth]{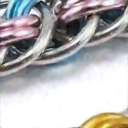}}
    \hfil
    \subfloat{\includegraphics[width=0.15\linewidth]{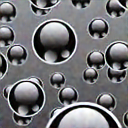}}
    \hfil
    \subfloat{\includegraphics[width=0.15\linewidth]{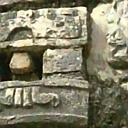}}
    \hfil
    \subfloat{\includegraphics[width=0.15\linewidth]{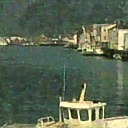}}
    \begin{center}
        \footnotesize  ~~~~~(a) Intra-dataset: ImageNet-1K-set  \qquad\qquad\qquad~~~~~(b) Inter-dataset: ImageNet-1K$\to$DTD\qquad\qquad~~ (c) Inter-dataset: ImageNet-1K$\to$BSDS500~~~~   
    \end{center}
    \caption{Visualizations of testing results on different datasets. The first row displays the noisy images with a Gaussian noise level of $\sigma = 0.25$. The second row showcases the denoised images generated by the PudNet using a single forward propagation.}
\label{denoised_img}
\vspace{-0.1in}
\end{figure*}

\subsection{Image Denoising Experiments}
We additionally conduct experiments on a typical image denoising task in the image processing field. Here, we leverage PudNet to generate parameters for a well-established convolutional blind-spot network \cite{krull2019noise2void}, which is built upon the UNet \cite{ronneberger2015u} architecture.

The noisy datasets are constructed based on both the intra-dataset setting and the inter-dataset setting. Following \cite{krull2019noise2void,xie2020noise2same}, noisy versions of all images are created by adding zero mean Gaussian noise with standard deviations of $\sigma =\{ 0.25,0.50,0.75 \}$.
We construct 20,000 sub-datasets comprising noisy images from the 800 classes in ImageNet-1K to train PudNet. For evaluating the performance of PudNet, in the intra-dataset setting, we utilize the testing noisy data from the remaining 200 classes in ImageNet-1K. 
Furthermore, we introduce the inter-dataset setting by constructing noisy cross-domain datasets. The noisy images from Animals-10, CIFAR-10, DTD, CIFAR-100, CUB-200 \cite{wah2011caltech}, and BSDS500 \cite{arbelaez2010contour} are used for evaluation. These six constructed cross-domain datasets are denoted as ImageNet-1K$\to$Animals-10,  ImageNet-1K$\to$CIFAR-10,  ImageNet-1K$\to$DTD,  ImageNet-1K$\to$CIFAR-100,  ImageNet-1K$\to$CUB-200,  and ImageNet-1K$\to$BSDS500.
To assess the performance of the image denoising, we use two widely-used metrics: peak signal-to-noise ratio (PSNR) \cite{hore2010image} and structural similarity (SSIM) \cite{wang2004image}.

\textbf{Performance Analysis of Intra-dataset Image Denoising.}
Table \ref{denoisy_intra} presents the results of image denoising in the intra-dataset setting. To demonstrate the time our method could save, we provide the time of training the blind-spot network from scratch using the widely-used optimizer Adam and the training acceleration technique, GC. 
We find that it takes around 100, 50, and 50 epochs to train the blind-spot network using the accelerated method GC and the network obtains PSNR values of 28.64 dB, 25.95 dB, and 24.64 dB, as well as SSIM values of 0.953, 0.919, and 0.892 for three different types of Gaussian noise, respectively.
In contrast, our method costs less than 1 GPU second to predict the parameters of the blind-spot network while still achieving the comparable performance (PSNR of 28.53 dB, 25.97 dB, and 24.52 dB, and SSIM of 0.950, 0.921, and 0.892) on the three datasets respectively, at least $6,000$ times faster than the training accelerated method.

\textbf{Performance Analysis of Inter-dataset Image Denoising.}
We further evaluate the performance of our PudNet with the image denoising task on six cross-domain datasets. We leverage PudNet trained on ImageNet-1K-set to directly generate parameters of the blind-spot network for six different cross-domain datasets. The results are reported in Table \ref{denoisy_inter1} and Table \ref{denoisy_inter2}, which show the efficiency of our PudNet. For instance, on the ImageNet-1K$\to$DTD dataset shown in Table \ref{denoisy_inter1}, 
our PudNet achieves comparable PSNR and SSIM metrics with that of training the model from scratch on DTD for over 100 epochs, while PudNet is at least 4,000 times faster than traditional training methods utilizing GC. 
On the ImageNet-1K$\to$BSDS500 dataset shown in the Table \ref{denoisy_inter2},  our PudNet achieves comparable PSNR and SSIM metrics with that of training the model from scratch on BSDS500 at around 300 epochs with Gaussian noise type ($\sigma=0.75$), while PudNet is at least 7,000 times faster than traditional training methods with GC. 

\textbf{Visualization of the Denoised Image Data.}
We further investigate the performance of PudNet in image denoising through visualizations. 
We randomly select two samples from ImageNet-1K-set in the intra-dataset setting, as well as two samples from ImageNet-1K$\to$DTD and ImageNet-1K$\to$BSDS500 in the inter-dataset setting respectively.
We leverage the PudNet trained on ImageNet-1K-set to generate parameters of the blind-spot network for these noisy images. As presented in Figure \ref{denoised_img}, our PudNet exhibits relatively satisfactory performance in generating network parameters for the image denoising task.

\section{Conclusion and Future Works}
In this paper, we identify correlations among image datasets and the corresponding parameters of a given ConvNet and explore a new training paradigm for ConvNets.
We propose a new hypernetwork, called PudNet, which could directly predict the network parameters for an image unseen dataset with only a single forward propagation.
In addition, we attempt to capture the relations among parameters across different network layers through a series of adaptive hyper-recurrent units. 
Extensive experimental results demonstrate the effectiveness and efficiency of our method in both intra-dataset prediction settings and inter-dataset prediction settings.

Nevertheless, the current work primarily delves into parameter generation for target networks based on CNNs. 
A potential avenue for future research could involve extending parameter generation to large language models (LLMs) \cite{touvron2023llama} or vision language models (VLMs) \cite{zhu2023minigpt}. 
For example, LLMs or VLMs commonly employ adapters \cite{chen2022vision} or LoRA methods \cite{hu2021lora} to enhance generalization to downstream tasks. This typically requires a certain amount of time for training. In future work, we can explore the potential of our approach in predicting parameters for adapters or LoRA methods, offering a pathway to save training time.



\bibliographystyle{IEEEtran}
\bibliography{tip_2023}

\begin{IEEEbiography}[{\includegraphics[width=1in,height=1.25in,clip,keepaspectratio]{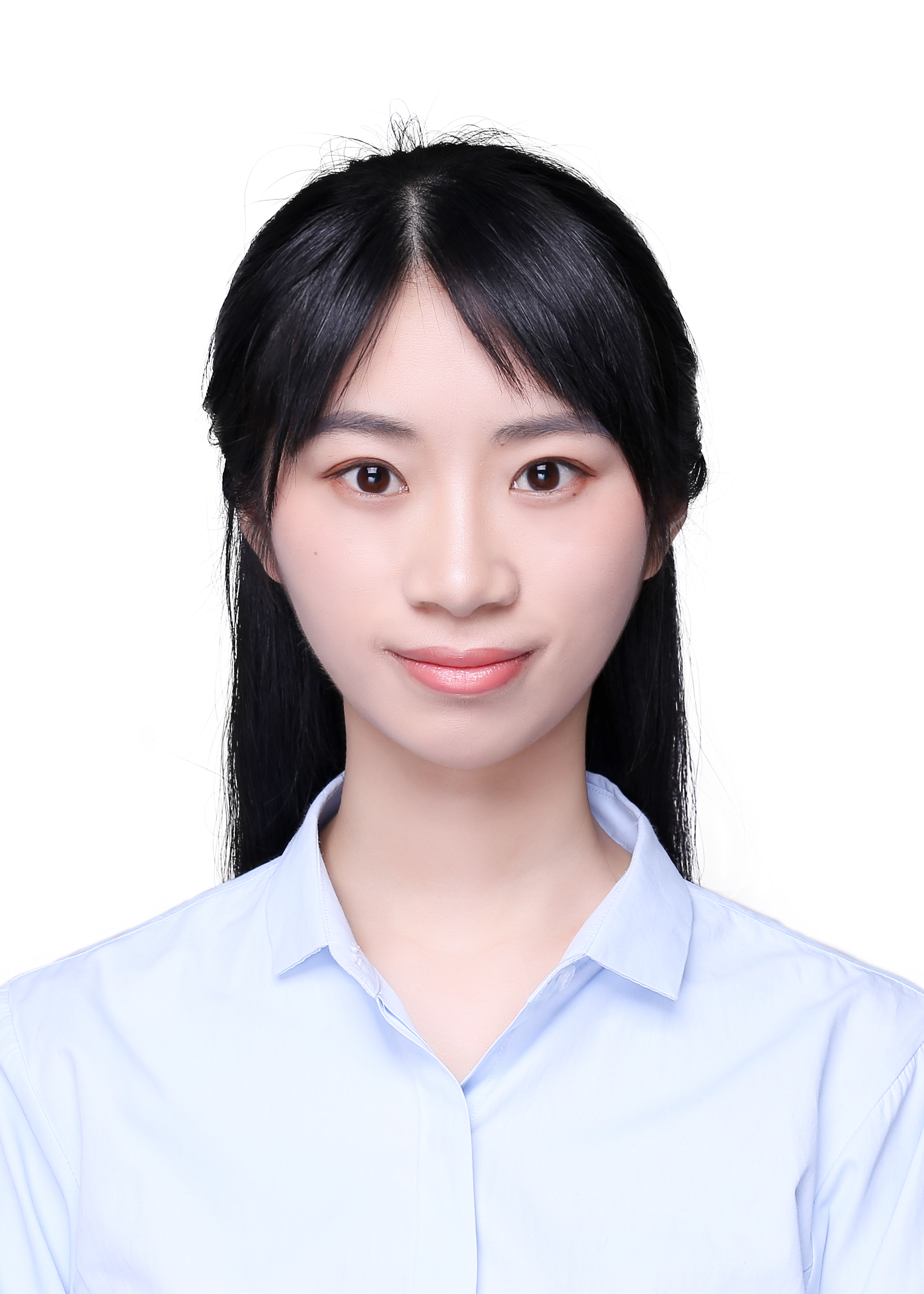}}]{Shiye Wang}
received the M.S. degree in computer science and technology from the Northeastern University (NEU) in 2020. She is currently pursuing the Ph.D. degree in computer science and technology from the Beijing Institute of Technology (BIT). Her research interests include multi-view clustering, unsupervised learning, and data mining.
\end{IEEEbiography}

\begin{IEEEbiography}[{\includegraphics[width=1in,height=1.25in,clip,keepaspectratio]{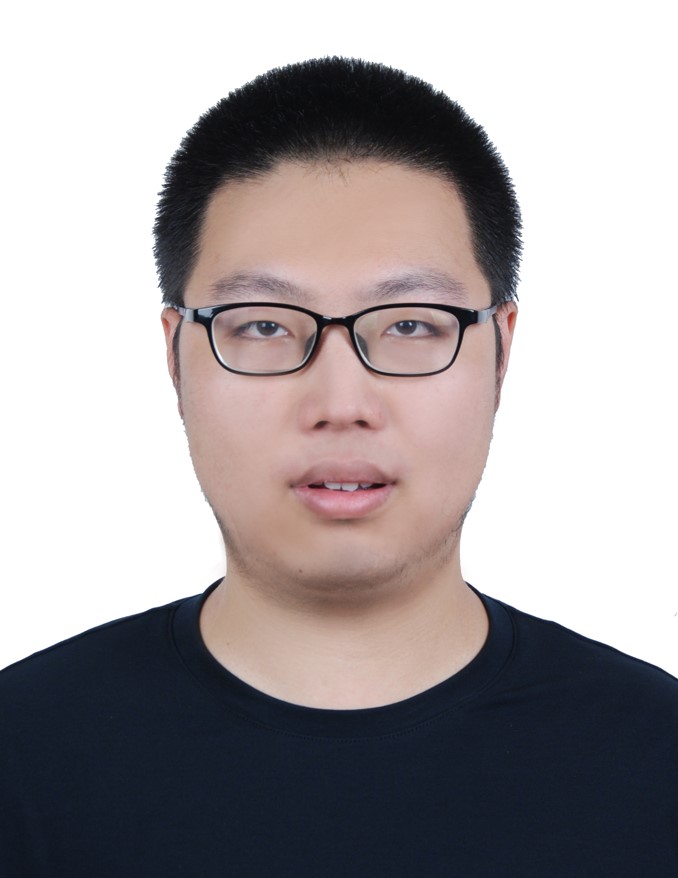}}]{Kaituo Feng}
received the B.E. degree in Computer Science and Technology from Beijing
Institute of Technology (BIT) in 2022. He is currently pursuing the master degree in Computer Science and Technology at Beijing Institute of Technology (BIT). His research interests include graph neural networks and knowledge distillation.
\end{IEEEbiography}

\begin{IEEEbiography}[{\includegraphics[width=1in,height=1.25in,clip,keepaspectratio]{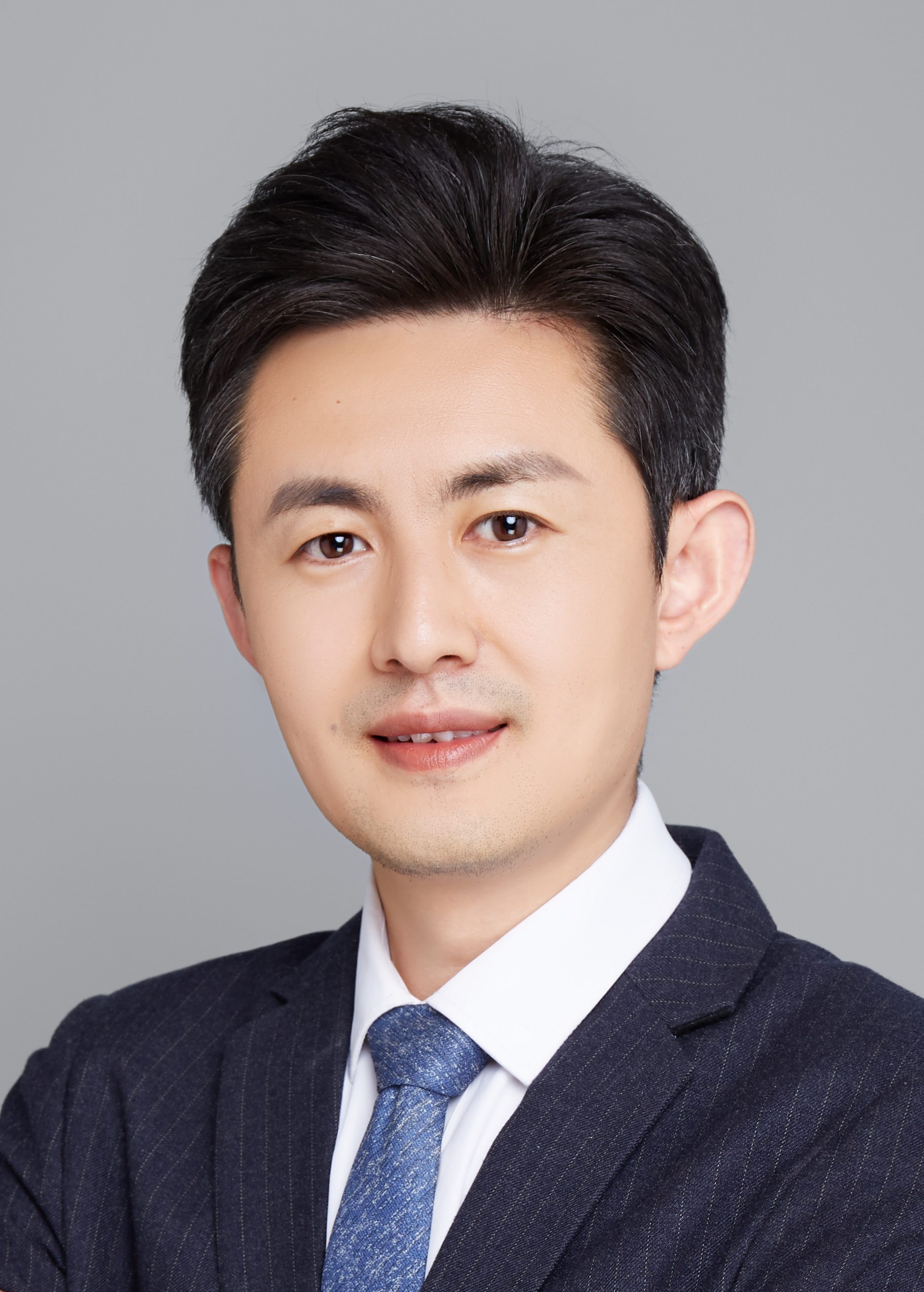}}]{Chengsheng Li}
received the B.E. degree from the University of Electronic Science and Technology of China (UESTC) in 2008 and the Ph.D. degree in pattern recognition and intelligent system from the Institute of Automation, Chinese Academy of Sciences, in 2013. During his Ph.D., he once studied as a Research Assistant with The Hong Kong Polytechnic University from 2009 to 2010. He is currently a Professor with the Beijing Institute of Technology. Before joining the Beijing Institute of Technology, he worked with IBM Research, China, Alibaba Group, and UESTC. He has more than 70 refereed publications in international journals and conferences, including IEEE TRANSACTIONS ON PATTERN ANALYSIS AND MACHINE INTELLIGENCE, IEEE TRANSACTIONS ON IMAGE PROCESSING, IEEE TRANSACTIONS ON NEURAL NETWORKS AND LEARNING SYSTEMS, IEEE TRANSACTIONS ON COMPUTERS, IEEE TRANSACTIONS ON MULTIMEDIA, NeurIPS, ICLR, ICML, PR, CVPR, AAAI, IJCAI, CIKM, MM, and ICMR. His research interests include machine learning, data mining, and computer vision. He won the National Science Fund for Excellent Young Scholars in 2021.
\end{IEEEbiography}

\begin{IEEEbiography}[{\includegraphics[width=1in,height=1.25in,clip,keepaspectratio]{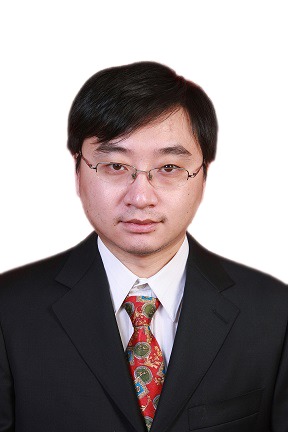}}]{Ye Yuan}
received the B.S., M.S., and Ph.D. degrees in computer science from Northeastern University in 2004, 2007, and 2011, respectively. He is currently a Professor with the Department of Computer Science, Beijing Institute of Technology, China. He has more than 100 refereed publications in international journals and conferences, including VLDBJ, IEEE TRANSACTIONS ON PARALLEL AND DISTRIBUTED SYSTEMS, IEEE TRANSACTIONS ON KNOWLEDGE AND DATA ENGINEERING,SIGMOD, PVLDB, ICDE, IJCAI, WWW, and KDD. His research interests include graph embedding, graph neural networks, and social network analysis. He won the National Science Fund for Excellent Young Scholars in 2016.
\end{IEEEbiography}

\begin{IEEEbiography}[{\includegraphics[width=1in,height=1.25in,clip,keepaspectratio]{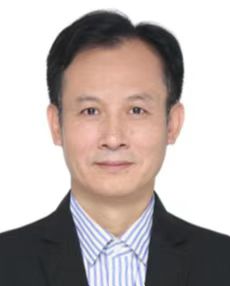}}]{Guoren Wang}
received the B.S., M.S., and Ph.D. degrees in computer science from Northeastern University, Shenyang, in 1988, 1991, and 1996, respectively. He is currently a Professor with the School of Computer Science and Technology, Beijing Institute of Technology, Beijing, where he has been the Dean since 2020. He has more than 300 refereed publications in international journals and conferences, including VLDBJ, IEEE TRANS-ACTIONS ON PARALLEL AND DISTRIBUTED SYSTEMS, IEEE TRANSACTIONS ON KNOWLEDGE AND DATA ENGINEERING, SIGMOD, PVLDB, ICDE, SIGIR, IJCAI, WWW, and KDD. His research interests include data mining, database, machine learning, especially on high-dimensional indexing, parallel database, and machine learning systems. He won the National Science Fund for Distinguished Young Scholars in 2010 and was appointed as the Changjiang Distinguished Professor in 2011.
\end{IEEEbiography}

\end{document}